\documentclass[letterpaper, 10 pt, conference]{ieeeconf}  % Comment this line out if you need a4paper

\IEEEoverridecommandlockouts                              % This command is only needed if 
\usepackage{graphics} % for pdf, bitmapped graphics files
\usepackage{epsfig} % for postscript graphics files
\usepackage{mathptmx} % assumes new font selection scheme installed
\usepackage{times} % assumes new font selection scheme installed
\usepackage{amsmath} % assumes amsmath package installed
\usepackage{amssymb}  % assumes amsmath package installed
\usepackage{algorithm}
\usepackage[noend]{algpseudocode}
\usepackage{subcaption}
\usepackage{textcomp}

\makeatletter
\def\BState{\State\hskip-\ALG@thistlm}
\makeatother

\title{\LARGE \bf
Planning Brachistochrone Hip Trajectory for a Toe-Foot Bipedal Robot going Downstairs 
}

\author{Gaurav Bhardwaj$^{{1},{\dag},{*}}$, Utkarsh A. Mishra$^{2,{\dag}}$, N. Sukavanam$^{3}$ and  R. Balasubramanian$^{1}$% <-this % stops a space
\thanks{Part of the manuscript has been accepted for presentation at the RoAI 2020: International Conference on Robotics and Artificial Intelligence 2020, IIT Madras and will be published in the Proceedings by the Journal of Physics: Conference Series.}
\thanks{$^{1}$Gaurav Bhardwaj and R.Balasubramanian are with the Computer Science and Engineering Department, IIT Roorkee
        {\tt\small gbhardwaj@cs.iitr.ac.in}, {\tt\small balarfcs@iitr.ac.in}}%
\thanks{$^{2}$Utkarsh A. Mishra is with the Mechanical and Industrial Engineering Department, IIT Roorkee
        {\tt\small umishra@me.iitr.ac.in}}%
\thanks{$^{3}$N. Sukavanam is with the Mathematics Department, IIT Roorkee
        {\tt\small nsukvfma@iitr.ac.in}}%
\thanks{$^{\dag}$ These authors have contributed equally.}%
\thanks{$^{*}$ Corresponding Author}%
}

\begin{document}

\maketitle
\thispagestyle{empty}
\pagestyle{empty}

%%%%%%%%%%%%%%%%%%%%%%%%%%%%%%%%%%%%%%%%%%%%%%%%%%%%%%%%%%%%%%%%%%%%%%%%%%%%%%%%

\begin{abstract}
 
A novel efficient downstairs trajectory is proposed for a 9 link biped robot model with toe-foot. Brachistochrone is the fastest descent trajectory for a particle moving only under the influence of gravity. In most situations, while climbing downstairs, human hip also follow brachistochrone trajectory for a more responsive motion. Here, an adaptive trajectory planning algorithm is developed so that biped robots of varying link lengths, masses can climb down on varying staircase dimensions. We assume that the center of gravity (COG) of the biped concerned lies on the hip. Zero Moment Point (ZMP) based COG trajectory is considered and its stability is ensured. Cycloidal trajectory is considered for ankle of the swing leg. Parameters of both cycloid and brachistochrone depends on dimensions of staircase steps. Hence this paper can be broadly divided into 4 steps 1) Developing ZMP based brachistochrone trajectory for hip 2) Cycloidal trajectory planning for ankle by taking proper collision constraints 3) Solving Inverse kinematics using unsupervised artificial neural network (ANN) 4) Comparison between the proposed, a circular arc and a virtual slope based hip trajectory. The proposed algorithms have been implemented using MATLAB\textregistered.      
\end{abstract}

%%%%%%%%%%%%%%%%%%%%%%%%%%%%%%%%%%%%%%%%%%%%%%%%%%%%%%%%%%%%%%%%%%%%%%%%%%%%%%%%
\section{INTRODUCTION}

Humanoids or biped robots have very unique characteristics which enables them to work like human in many challenging situations as compared to wheeled robots. One of the important task is climbing on stairs, where a biped robot will be more useful to perform kind of activities currently performed by humans. However achieving stability in these types of environments is quite difficult in case of such robots due to their higher tendency to fall as compared to wheeled or quadruped robots. A lot of research has been done in past for making biped robots to climb stairs.

Shih and chio \cite{c1} studied static model for biped to walk on stairs. Kajita et al. \cite{c2} proposed preview control of ZMP and applied same on spiral staircase. Jeon et al. \cite{c3} proposed optimal trajectory generation method based on genetic algorithms to walk upstairs. Morisawa et al. \cite{c4} has given technique for pattern generation of biped walking constraint on parametric surface. Sato et al. \cite{c5} proposed a virtual slope method for staircase walking for biped robot.

Artificial neural networks(ANN) also played a vital role in robotics in past two decades. \cite{c9,c10} used neural network for inverse kinematics solution with training data obtained from relationship between joint coordinates and end effector Cartesian coordinates. In a previous research \cite{c13}, the authors have also proposed an unsupervised ANN technique for inverse kinematics solution for 6 degree of freedom PUMA (Programmable Universal Machine for Assembly) robot.

One of the most important aspect in case of biped robot is its stability which makes it to walk on flat or rough terrains without falling down. Vukobratovic \cite{c15} in 1972 proposed a measuring index for stability called Zero Moment Point. Foot Rotation Indicator(FRI) \cite{c16} , Contact Wrench sum and Contact Wrench Cone \cite{c17} are other indicators of stability in biped robots. ZMP is one of the simplest and most used technique which is used in this paper.

In this paper, a novel brachistochrone trajectory realization for hip is considered to perform downstairs motion by a toe-foot biped robot model with cycloidal ankle trajectory. The motion is supported by an adaptive planning strategy for the initial hip height based on the geometrical properties of the robot as well as the stairs. An appropriate comparison with a circular arc and a virtual slope based hip trajectory is drawn. The brachistochrone path is already a fastest descent path under the action of the gravitational force and hence the trajectory developed on such a path resembles human like motion more closely.

The following section describes the robot model, its geometrical properties and establishes the problem statement. Section III presents the trajectory planning strategy for the all the joints which is used in Section IV to get the joint space solutions needed to track the planned trajectory. Finally the complete simulation comparison of the various hip trajectories are presented and discussed in Section V. The work concludes in Section VI.

\section{ROBOT MODEL DESCRIPTION}

\begin{figure}[hp]
      \centering
      \includegraphics[scale=0.35]{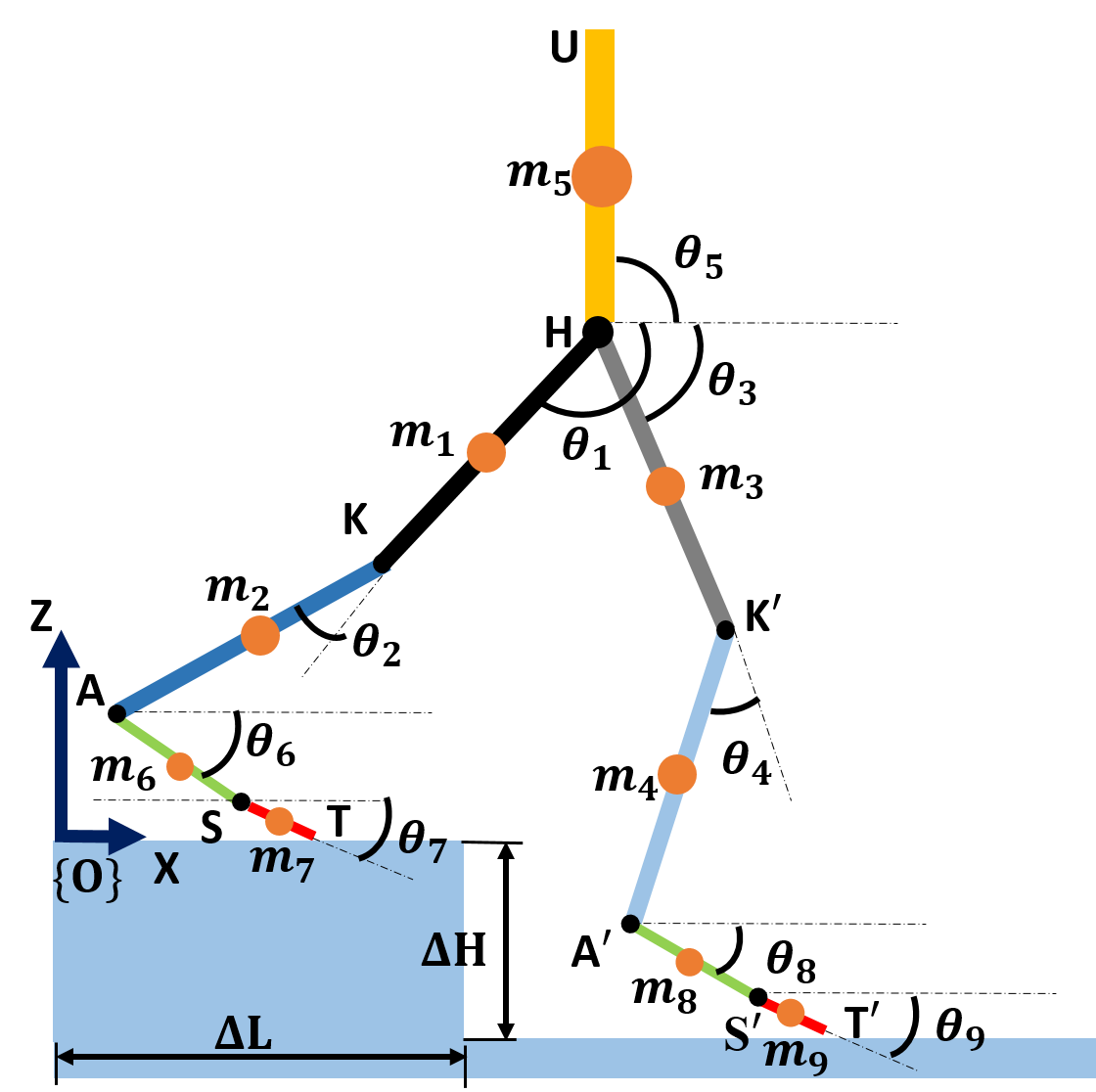}
      \caption{Toe-Foot Robot Model Description and Notations}
      \label{modelpic}
   \end{figure}

A toe-foot joint planar bipedal robot model is considered with it's complete movement constrained to the sagittal plane. The model consists of a upper body and two legs. Each leg consists of 4 links and 4 joints, all of which are revolute in nature as shown in Fig. \ref{modelpic}. These joints are referred as the Hip(H), Knee(K), Ankle(A), Sole(S) along with the Toe tip(T). The points in the figure corresponding to each of the joints are shown in Table \ref{tablemodel} and the corresponding attributes of each segment of the model can be visualized from both the Fig. \ref{modelpic} and Table \ref{tablemodel}.

\begin{table}[h]
\caption{Joint Positions and Attributes of Each Link}
\label{tablemodel}
\centering
\begin{tabular}{|c|c|c|c|c|c|}
\hline
\multicolumn{2}{|c}{Joint Name} & \multicolumn{4}{|c|}{Positions}\\
\hline
\multicolumn{2}{|c}{Hip(H)} & \multicolumn{4}{|c|}{H (both Swing \& Stance Leg)}\\
\hline
\multicolumn{2}{|c}{Knee(K)} & \multicolumn{4}{|c|}{K (Swing Leg), K$'$ (Stance Leg)}\\
\hline
\multicolumn{2}{|c}{Ankle(A)} & \multicolumn{4}{|c|}{A (Swing Leg), A$'$ (Stance Leg)}\\
\hline
\multicolumn{2}{|c}{Sole(S)} & \multicolumn{4}{|c|}{S (Swing Leg), S$'$ (Stance Leg)}\\
\hline
\multicolumn{2}{|c}{Toe(T)} & \multicolumn{4}{|c|}{T (Swing Leg), T$'$ (Stance Leg)}\\
\hline
\multicolumn{6}{|c|}{ } \\
\hline
Link No. & Link & Length & Value(cm) & Mass & Value(Kg)\\
\hline
1 & HK & $l_1$ & 40 & $m_1$ & 6\\
\hline
2 & KA & $l_2$ & 40 & $m_2$ & 4\\
\hline
3 & HK$'$ & $l_3$ & 40 & $m_3$ & 6\\
\hline
4 & K$'$A$'$ & $l_4$ & 40 & $m_4$ & 4\\
\hline
5 & UH & $l_5$ & 30  & $m_5$ & 30\\
\hline
6 & AS & $l_6$ & 12  & $m_6$ & 0.70\\
\hline
7 & ST & $l_7$ & 5 & $m_7$ & 0.15\\
\hline
8 & A$'$S$'$ & $l_8$ & 12 & $m_8$ & 0.70\\
\hline
9 & S$'$T$'$ & $l_9$ & 5 & $m_9$ & 0.15\\
\hline
\end{tabular}
\end{table}

So, the overall length of each leg can be given by $(l_1+l_2)$ and that of foot is given by $(l_6+l_7)$. The approach of modeling any gait is primarily done by considering hip as the base and and the ankle of the active leg as the end-effector. Such a 2-link manipulator configuration is then used to divide the movement into 2 phases, namely the Double Support Phase (DSP) and the Single Support Phase(SSP). The former one corresponds to the initial condition when both the feet are making contact with the ground whereas the latter represents the swing of the active leg with passive leg's foot remaining in contact with the ground.

The presented work explores the possibilities of stable downstairs climbing for the model shown, with known step length and height. The complete task objective is to enable the model to climb down a singles step as well as subsequent steps progressively.

\section{TRAJECTORY PLANNING}

\subsection{For Swing Leg}

A complete swing leg motion is achieved during the time interval $(t_0 = 0,t_f )$, which is divided majorly into four phases.

\subsubsection{Double Support Phase ($t = 0 \textbf{ to } t_2$)}

The foot of the swing leg denoted by points A, S and T are on the grounds with coordinates $(0,0)$, $(l_6 , 0)$ and $(l_6 + l_7 , 0)$ respectively. The feet is considered to move as a two-link manipulator with AS and ST as the two arms about the fixed toe, T.

Thus, during this phase, the ankle coordinates are given by the following equations.
\begin{gather}
x_A (t) = l_6 (1-cos(\theta_{6} (t))) + l_7 (1-cos (\theta_{7} (t)) \\
z_A (t) = l_6 sin(\theta_{6} (t)) + l_7 sin (\theta_{7} (t))
\end{gather}

For the duration, the sole AS rotates about S, from angle $0$ at $t = 0$ to angle $\theta_a$ at time $t = t_1$ with toe being stable on ground. For the time duration, $(t_1 , t_2 )$ the AS-ST links behave as 2 link manipulator about T where AS reverses its motion and ST rotates from angle $0$ at $t = t_1$ to angle $\theta_b$ at time $t = t_2$ . The parameters $\theta_a$ and $\theta_b$ are modeling parameters for stable gait generation.

\subsubsection{Single Support Phase ($t = t_2 \textbf{ to } t_f$)}

A cycloidal path was used to model the swing leg motion of the ankle. Such a  planning enables null accelerations at the end of the gait and hence, the dynamic forces on the links are not significant for that instant. The trajectory parameters used for the study is based on the robot's geometrical as well as the stair properties.

Consider a cycloid of constant radius $r$ and a time parametrized $\theta (t)$, on which the Cartesian coordinates of a point is given by,
\begin{gather}
x_{cycloid}(t) = r (\theta_c(t) - \sin{\theta_c(t)}) \\
z_{cycloid}(t) = r (1 - \cos{\theta_c(t)})
\end{gather}
The maximum height that can be achieved in such a motion is $2r$ at $\theta=\pi$ which is at a horizontal distance of $\pi r$. Now, to climb down a step of height ($\Delta z$) and at certain distance ($\Delta x$), $r$ is considered to be equal to $\Delta z/2$ and cycloid is formed with the position of the ankle $(x_A(t_f),z_A (t_f))$, at the beginning of the DSP of the stance leg, as its end position. The cycloid is governed by the value of $\theta_c = 2\pi$ at the end and back-traced till $\theta_c = \theta_{c0}$. After the DSP ends, the remaining distance to be covered by the trajectory happens to be $\Delta x_c = (\Delta x - x_A(t_2))$. The value of $\theta_c$ in the beginning, $\theta_{c0}$, is chosen such that it satisfies,

\begin{equation}
 r (\theta_{c0} - \sin{\theta_{c0}}) =
\begin{cases}
    \pi r - \Delta x_c,& \text{if } \pi r \geq \Delta x_c\\
    0,              & \text{otherwise}
\end{cases}
\end{equation}
Also, (5) shows the dependence of the value of $x_{c0}$ on $r$ and $\Delta x_c$. As the DSP is bounded by the workspace and feasibility constraints of a 2-link mechanism, it becomes difficult for it to follow the above described cycloidal trajectory just after it ends and hence a bridge between them is modeled as a bezier curve. The trajectory was formulated based on 4 control points which starts from the end of DSP of swing leg. The second control point was chosen based on the cycloid formulation, with the value of $\theta_{c0}$.The remaining control points were selected from within the cycloid such that the bezier curve blends completely with the cycloidal trajectory. Now, to maintain the continuity the velocity at the blend of the bezier and cycloid curves were set to be same.

Also, to make the complete utilization of cycloidal trajectory, $\ddot{z}_c(\theta_c = \pi) = g$, where $g$ is the acceleration due to gravity. The complete cycloid definition can be completed by defining $\theta_c(t)$ as a polynomial parametrization based on all the boundary conditions.

\subsubsection{Movement of Sole and Toe}

The trajectory of the sole (S) and toe (T) during the complete motion $(0 \leq t \leq t_f)$ were chosen based on the geometry of the robot and the stair-step with which it is interacting. It was assured that no collision with the surface occurs at any point.

Once, the swing leg lands with it's toe(T) touching the stair, the next DSP phase starts and the ankle follows the exact reverse trajectory of that discussed in the first part of this segment i.e. first as a 2-link manipulator with fixed axis at T and then a single link manipulator with fixed axis at sole(S).

\subsection{For Hip Motion}

The hip is taken to be the moving base of the planar biped model. Hence, the movement of the hip plays a very vital role in governing the overall behavior of the gait. The motion of the hip was formulated as a COG trajectory with the ZMP equation of a one-mass COG model in sagittal plane. During the single support phase, the analytical solution of x-ZMP trajectory is considered as the x-COG trajectory which is directly in correspondence to the hip motion in x-direction, assuming the torso to be straight upward i.e. $\theta_5 = \pi/2$ always. Additionally, the ZMP motion is planned as a 3-degree polynomial parametrization with C-2 continuity. For such a formulation, the ZMP trajectory in saggital plane is as follows.
\begin{gather}
p_x(t) = x_C(t) - \frac{z_C}{g} \ddot{x}_C(t) \\
p_x(t) = a_{z0} + a_{z1} t + a_{z2} t^2 + a_{z3} t^3
\end{gather}
where $(p_x(t),0)$ and $(x_C,z_C)$ represent the ZMP and COG position respectively. The analytical solution for the ZMP equations (20) and (21), and the COG trajectories is obtained as follows,
\begin{gather}
\begin{split}
x_C(t) = C_1 e^{\omega t} + C_2 e^{-\omega t} + a_{z0} + a_{z1} t + a_{z2} t^2 + a_{z3} t^3 \\
+ \frac{z_C}{g}(6 a_{z3} t + 2 a_{z2})
\end{split} \\
\omega = \sqrt{\frac{g}{z_{Ci}}}
\end{gather}
where, $z_{Ci}$ is the initial centroidal height. The nature of behavior an individual adapts during climbing down the stairs is quite dynamic as for a fixed step height and length, individuals with varied height can comfortably  execute the motion. Such a plan can be loosely referred as an adaptive planning strategy where the individual lowers the hip to the extent that the toe of the stance leg just touches the next stair. An inverse planning is done to obtain the required initial height given the toe of the stance leg as the fixed base with all angles of the leg (a 4-link serial manipulator) set to maximum and constraint to the x-coordinate of the hip based on the one mass COG model.

\begin{figure}[thpb]
      \centering
      \includegraphics[scale=0.28]{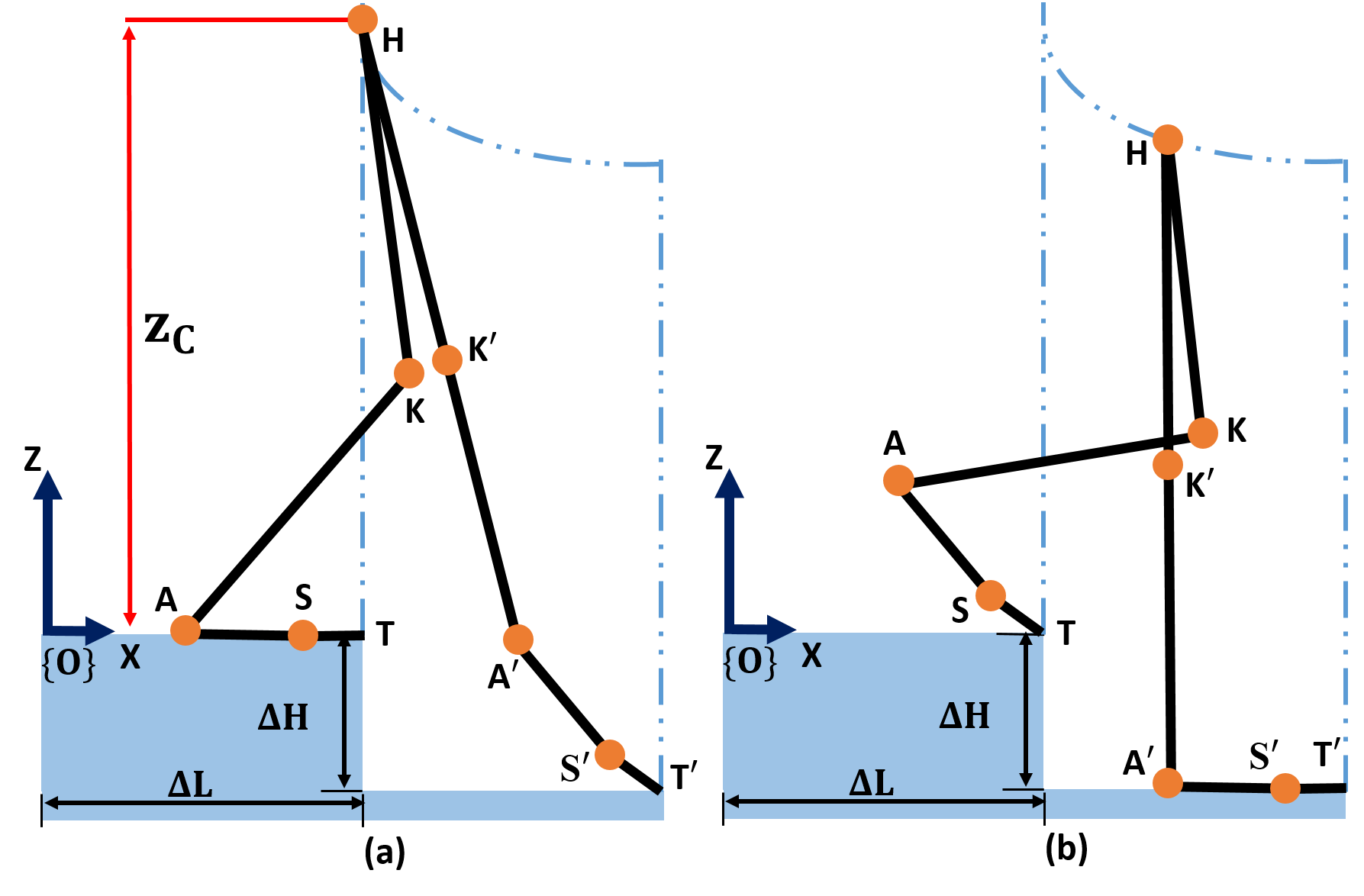}
      \caption{Adpative Hip Motion Behaviour (Torso not shown)}
      \label{transition}
   \end{figure}

Furthermore, climbing down the stairs should ideally take place with the natural application of the gravity force. In this respect, it is widely known that the brachistochrone curve is the fastest path. A comparison is done between various hip trajectories based on the same x-ZMP formulation. The hip motion in z-direction is modeled as s brachistochrone, then as a circular arc and finally according to the virtual slope method.

\subsubsection{Brachistochrone Formulation}

This curve traverses between the initial hip position $(x_H^{initial},z_H^{initial})$ to the final hip position $(x_H^{final},z_H^{final})$ in a more practical manner. Let $R_{H}$ and $\theta_{H}$ denote the segment corresponding to the desired brachistochrone arc, where the brachistochrone with a fixed radius $R_{H}$ starts from $\theta_{H}$ and continues to $\pi$.
\begin{equation}
\begin{gathered}
R_H (\pi - \theta_H + \sin(\theta_H)) = x_H^{final} - x_H^{initial} \\
R_H (2-1+\cos(\theta_H)) = z_H^{final} - z_H^{initial}
\end{gathered}
\end{equation}
The above pair of equations can be solved to get the corresponding values of $R_{H}$ and $\theta_{H}$. The hip z-coordinate $z_b$  as a function of the hip x-coordinate $x_b$ based on one mass COG model can be given as obtaining $\theta_{b}$ by solving,
\begin{equation}
x_H^{final} - \pi R_{H} + R_{H} (\theta_{b} - \sin \theta_{b}) = x_b
\end{equation}
and then using the obtained value of $\theta_{b}$ to get the corresponding $z_b$ according to,
\begin{equation}
z_b = z_H^{final} - 2 R_{H} + R_{H} (1 - \cos \theta_{b})
\end{equation}
It should be noted that this is irrespective of the x-zmp trajectory used and depicts a brachistochrone path as a fastest descent segment.

\subsubsection{Circular Arc Formulation}

Similar to the brachistochrone curve, his curve traverses between the initial hip position $(x_H^{initial},z_H^{initial})$ to the final hip position $(x_H^{final},z_H^{final})$ in the form of a circular arc. Let radius,$R_{H}$, and $\theta_{H}$ denote the segment of the circle, then,
\begin{equation}
\begin{gathered}
R_H \sin(\theta_H) = x_H^{final} - x_H^{initial} \\
R_H (1 - \cos(\theta_H)) = z_H^{final} - z_H^{initial}
\end{gathered}
\end{equation}
These equations are solved to get the values of $R_{H}$ and $\theta_{H}$, which were further used to obtain the hip coordinates by solving similarly for $\theta_b$ from,
\begin{equation}
x_H^{final} - R_H \sin(\theta_H - \theta_b) = x_b 
\end{equation}
and using it to obtain $z_b$ as,
\begin{equation}
z_b = z_H^{final} - R_{H} (1 + \cos (\theta_H - \theta_{b}))
\end{equation}

\subsubsection{Virtual Slope Formulation}

A simple straight line path connecting the initial and final hip positions is formulated according to the virtual slope, $k$, obtained from the positions. Thus,
\begin{equation}
k = \frac{z_H^{final} - z_H^{initial}}{x_H^{final} - x_H^{initial}}
\end{equation}
and accordingly, for a particular $x_b$, the corresponding $z_b$ can be obtained as,
\begin{equation}
z_b = z_H^{initial} + k (x_b - x_H^{initial})
\end{equation}

All the above formulations were calculated based on the value of coefficients $a_{zi}'s$, $C_1$ and $C_2$ in equation (23) which is determined using the boundary conditions of ZMP and COG respectively for the time-interval $(0, t_3)$. These boundary conditions are given in the following equations.
\begin{gather}
\begin{gathered}
p_x(t=0) = x_{ZMP}^{initial}, \; p_x(t=t_{stop}) = x_{ZMP}^{final}\\
\dot{p}_x(t=0) = 0, \; \dot{p}_x(t=t_{stop}) = 0 \\
x_C(t=0) = x_{COG}^{initial}, \; x_C(t=t_{stop}) = x_{COG}^{final}\\
\end{gathered}
\end{gather}

The above equation implies the transition from Fig. \ref{transition} part (a) to (b) where the transition starts from (a) at $t=0$ to (b) at $t=t_2$ and  moves to again (a) relative to the next stair until $t=t_{stop}$. Here $t=t_{stop}$ is a little before $t=t_f$ when the complete gait finishes.

All the transition parameters are characteristics of the behavior of the model during the gait as the model has to execute the hip trajectory as a moving base such that desired ankle trajectories (discussed in the previous subsection) can be tracked within the feasible workspace of the manipulator.

\subsection{For Stance leg}

The stance leg follows the hip trajectory while keeping the ankle fixed. Furthermore, the sole (S) and toe (T) of stance leg remain at the same point. As the swing leg starts DSP for the next phase, the initial-DSP for stance leg commences.

\section{INVERSE KINEMATICS}

\subsection{Inverse Kinematics formulation}
Forward kinematics equations were derived from \cite{c13} based on D-H Procedure. The previous trajectory planning segment clearly defines the trajectory of the hip (H), ankle (A), sole (S) and toe (T). Considering hip (H) as the base and ankle (A) as the end-effector, the forward kinematics equation is obtained as a function of the joint angles $\theta_1(t)$ and $\theta_2(t)$.
\begin{equation}
x_A(t) = x_H(t) + (l_1 \cos{\theta_1(t)} + l_2 \cos(\theta_1(t)+\theta_2(t)))
\end{equation}
\begin{equation}
z_A(t) = z_H(t) - (l_1 \sin{\theta_1(t)} + l_2 \sin(\theta_1(t)+\theta_2(t)))
\end{equation}
Here, (26) and (27) apply for both the swing leg and the stance leg with joint angles $\theta_1(t)$, $\theta_2(t)$ and $\theta_3(t)$, $\theta_4(t)$ respectively.

\subsection{Artificial Neural Network (ANN) approach}

An ANN based approach was used to solve the inverse kinematics problem from the forward kinematics equations (26) and (27). A feed forward network was modeled and trained in real time for every time instant. The hyper-parameters of the network are given in Table \ref{tableiknn}. 
\begin{table}[h]
\caption{Hyper-parameters for Feed Forward Network}
\label{tableiknn}
\begin{center}
\begin{tabular}{|c|c|}
\hline
Parameter & Value\\
\hline
Input Neurons & 2\\
\hline
Output Neurons & 2\\
\hline
Hidden layer & 1\\
\hline
Hidden layer nodes & 10\\
\hline
Activation function & Sigmoid $\frac{1}{1+e^{-x}}$\\
\hline
Learning Rate & $10^{-4}$\\
\hline
Maximum Iterations & 5000\\
\hline
\end{tabular}
\end{center}
\end{table}
For a desired ankle position $(x_A^{input},z_A^{input})$, the joint angles $\theta_1^{network}$ and $\theta_2^{network}$ were approximated. These approximated joint angles were substituted in (26) and (27) to obtain the network estimated ankle position $(x_A^{network},z_A^{network})$. Finally, the loss (error) function, $E_{network}$, was calculated as the squared error between the desired and approximated ankle positions.
\begin{equation}
E_{network} = (x_A^{input}-x_A^{network})^2 + (z_A^{input}-z_A^{network})^2
\end{equation}
The loss function was optimized using a gradient descent based on the partial derivative of the error with respect to the layer weights. Finally, the weights$(W)$ were updated according to the calculated gradients $(\delta)$, for each time step i.e.
\begin{equation}
W_{n+1}(t) = W_n(t) - \alpha \delta ,
\end{equation}
where $n$ represents the iteration (epoch) number. The weight updates were stopped once the error goes down below the threshold equal to $10^{-6}$.

\section{RESULT AND DISCUSSIONS}

A complete downstairs climbing methodology is been presented starting from the first step to climbing subsequent steps gradually. This section compares the simulation results obtained by the various hip trajectories formulated based on the same x-ZMP motion for one mass COG model. The various governing parameters of the presented trajectory planning scheme along with their considered values are shown in Table \ref{table:param}. These parameters are invariant and used for all the simulations. 

   \begin{table}[h]
\caption{Considered Value of Trajectory Governing Parameters}
\label{table:param}
\centering
\begin{tabular}{|c|c|c|c|}
\hline
Parameter & Value & Parameter & Value\\
\hline
$t_1$ & 1.00 sec & $t_3$ & 3.50 sec\\
\hline
$t_p$ & 1.25 sec & $\theta_a$ & $\pi/3$ rad\\
\hline
$t_2$ & 1.50 sec  & $\theta_b$ & $\pi/3$ rad\\
\hline
$t_{stop}$ & 3.00 sec & $\theta_{c0}$ & $\pi$ + 0.01 rad\\
\hline
\end{tabular}
\end{table}

\begin{figure}[thpb]
      \centering
      \includegraphics[scale=0.40]{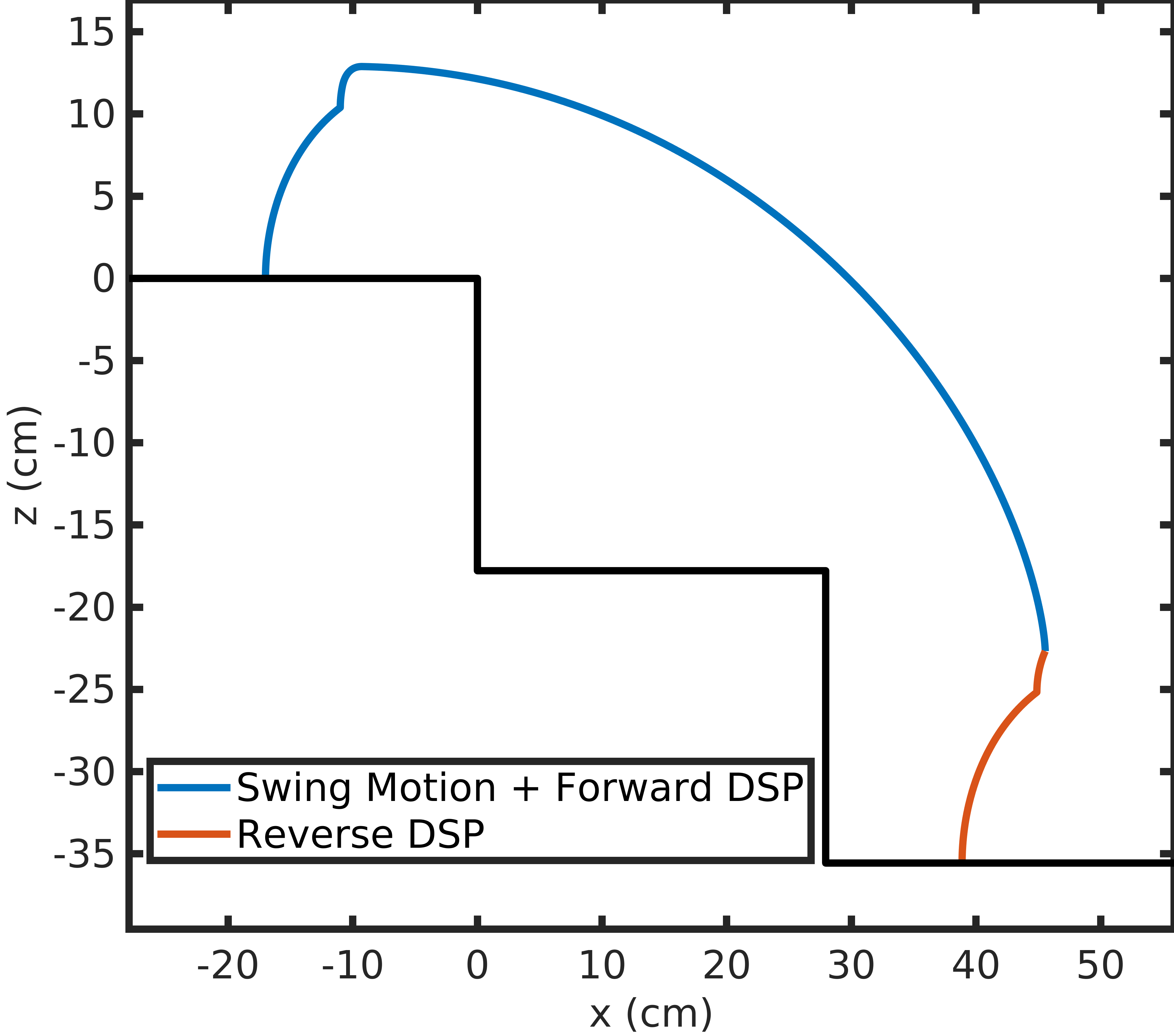}
      \caption{Ankle Trajectory of Swing Leg}
      \label{fig:anktraj}
   \end{figure}

Based on the values of the parameters, the swing leg ankle trajectory was formulated as a combination of the DSP, the spline bridge and the cycloidal realization as shown in Fig.  \ref{fig:anktraj}. This trajectory of the ankle is taken to be the same for all the simulations.

\begin{figure}[thpb]
      \centering
      \includegraphics[scale=0.40]{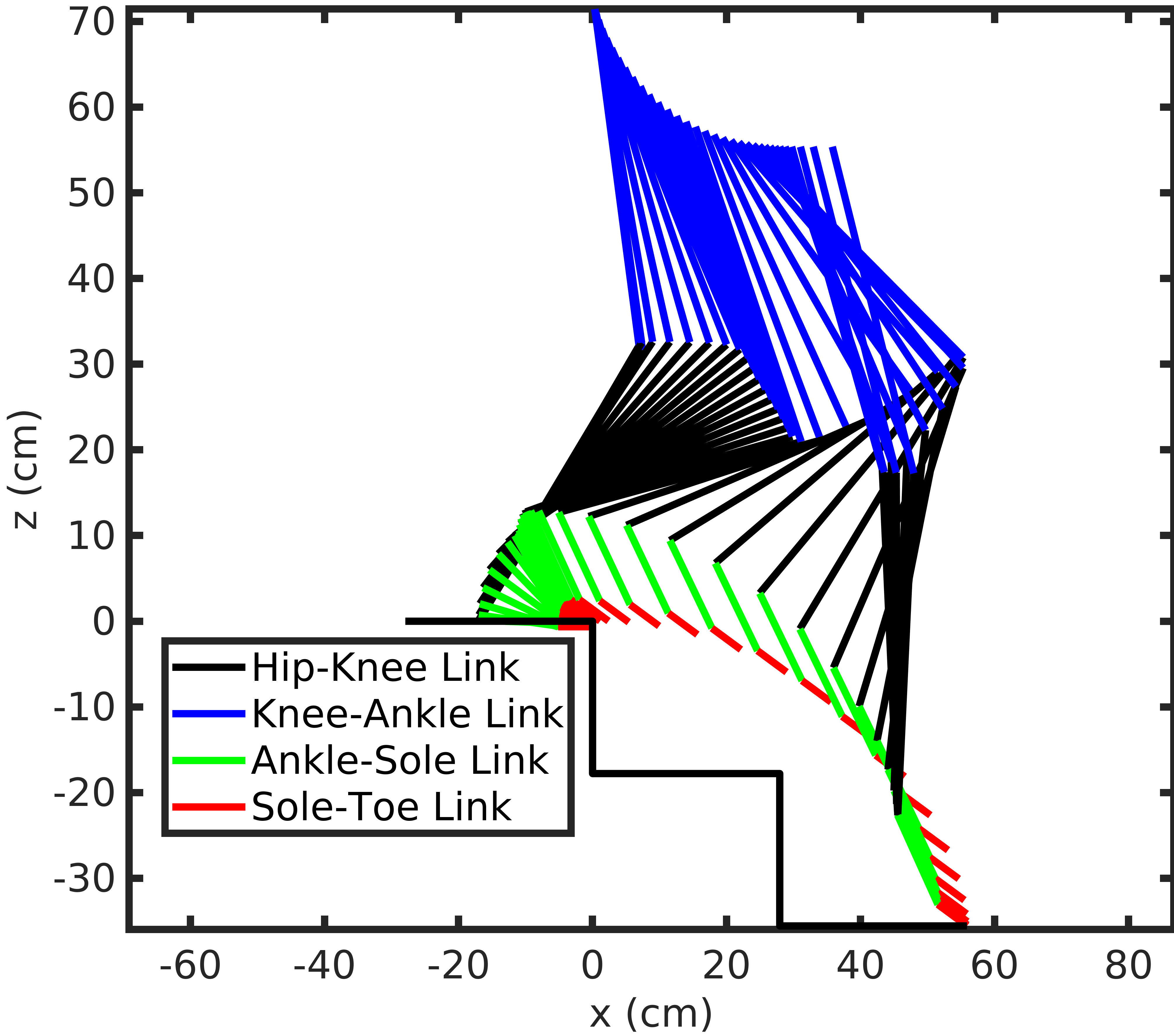}
      \caption{Complete Swing Leg Trajectory for Brachistochrone Hip trajectory}
      \label{fig:swing}
   \end{figure}

\begin{figure*}
\begin{subfigure}{.16\textwidth}
  \centering
  \includegraphics[width=\linewidth]{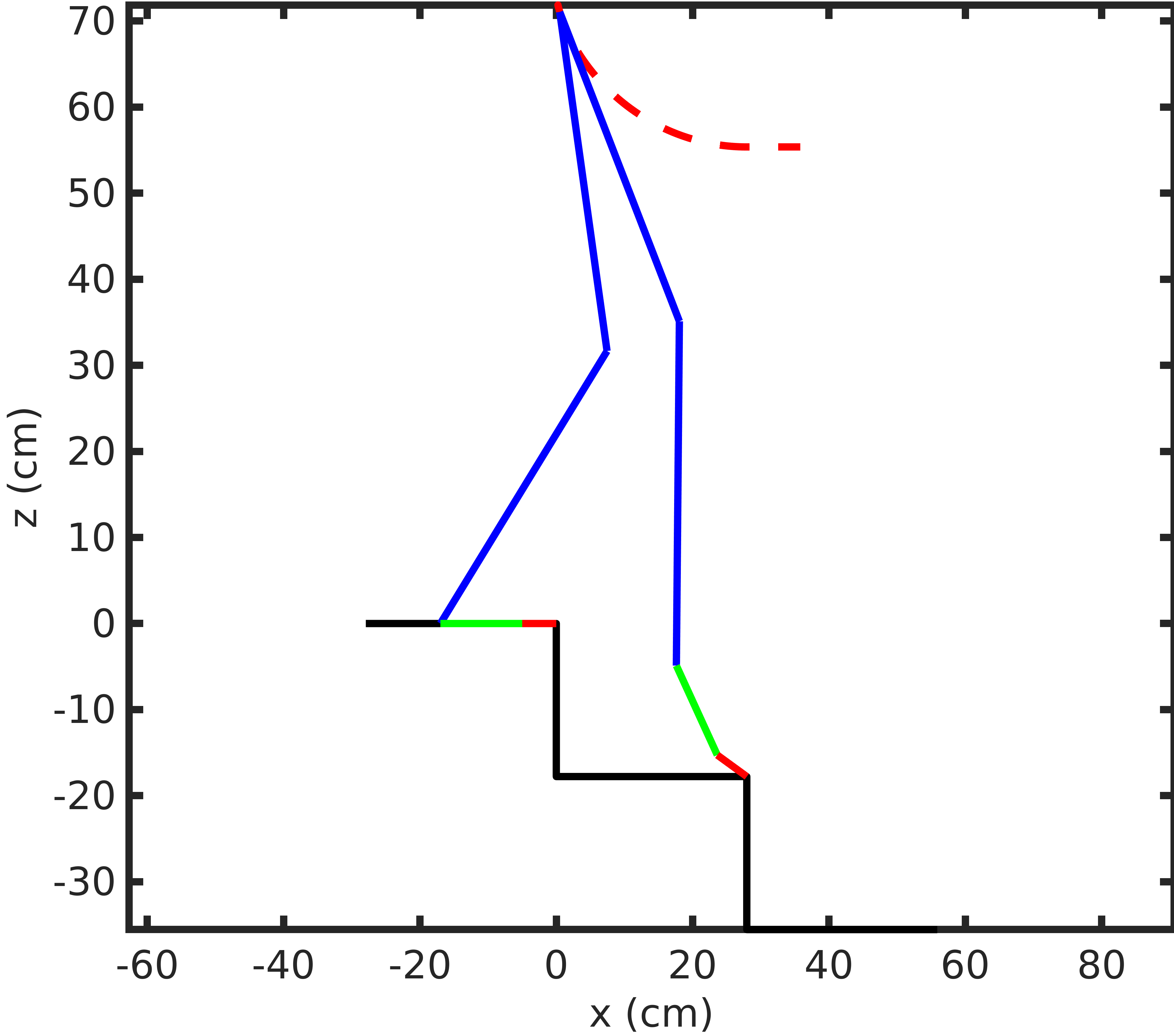}
\end{subfigure}%
\begin{subfigure}{.16\textwidth}
  \centering
  \includegraphics[width=\linewidth]{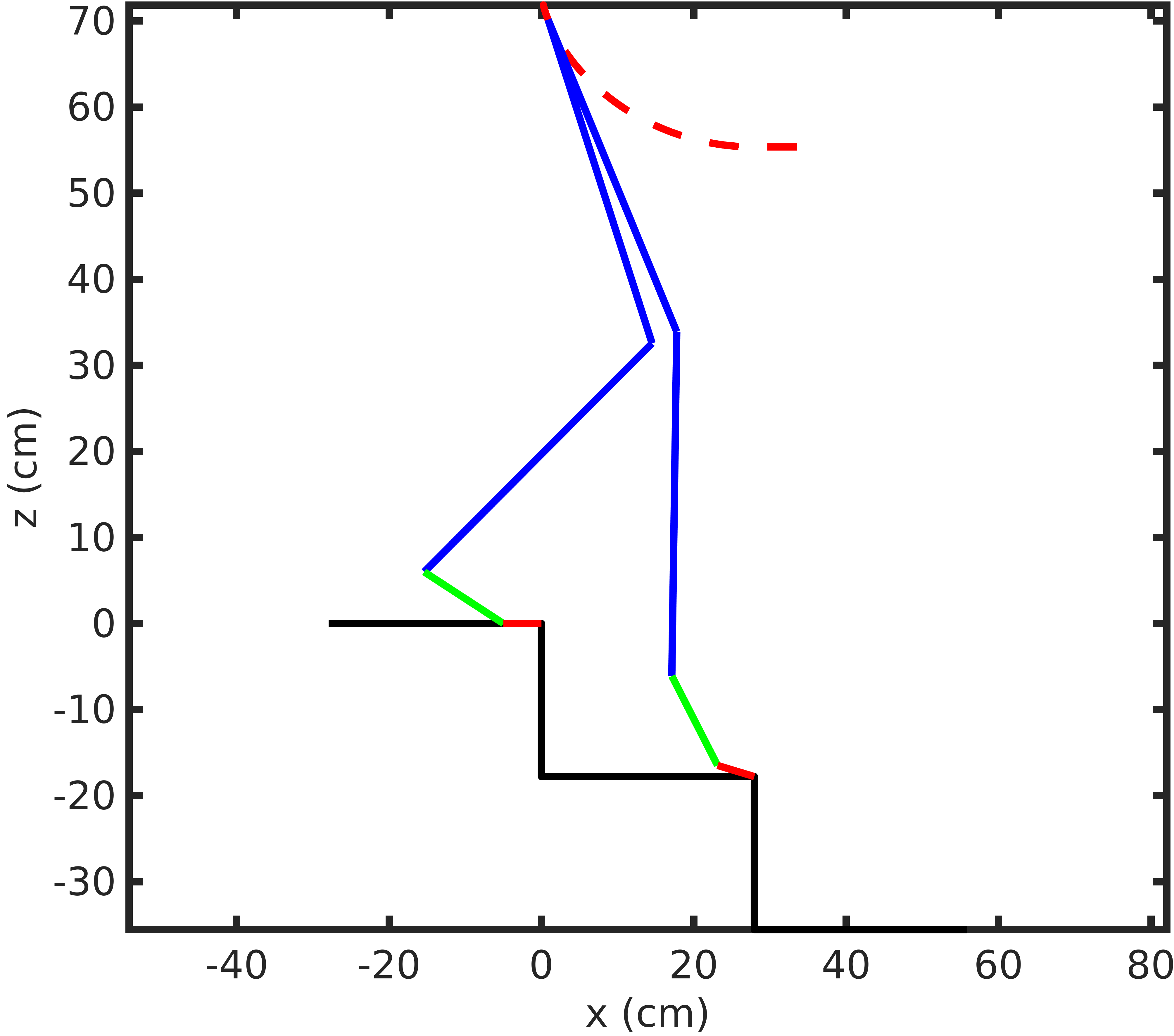}
\end{subfigure}
\begin{subfigure}{.16\textwidth}
  \centering
  \includegraphics[width=\linewidth]{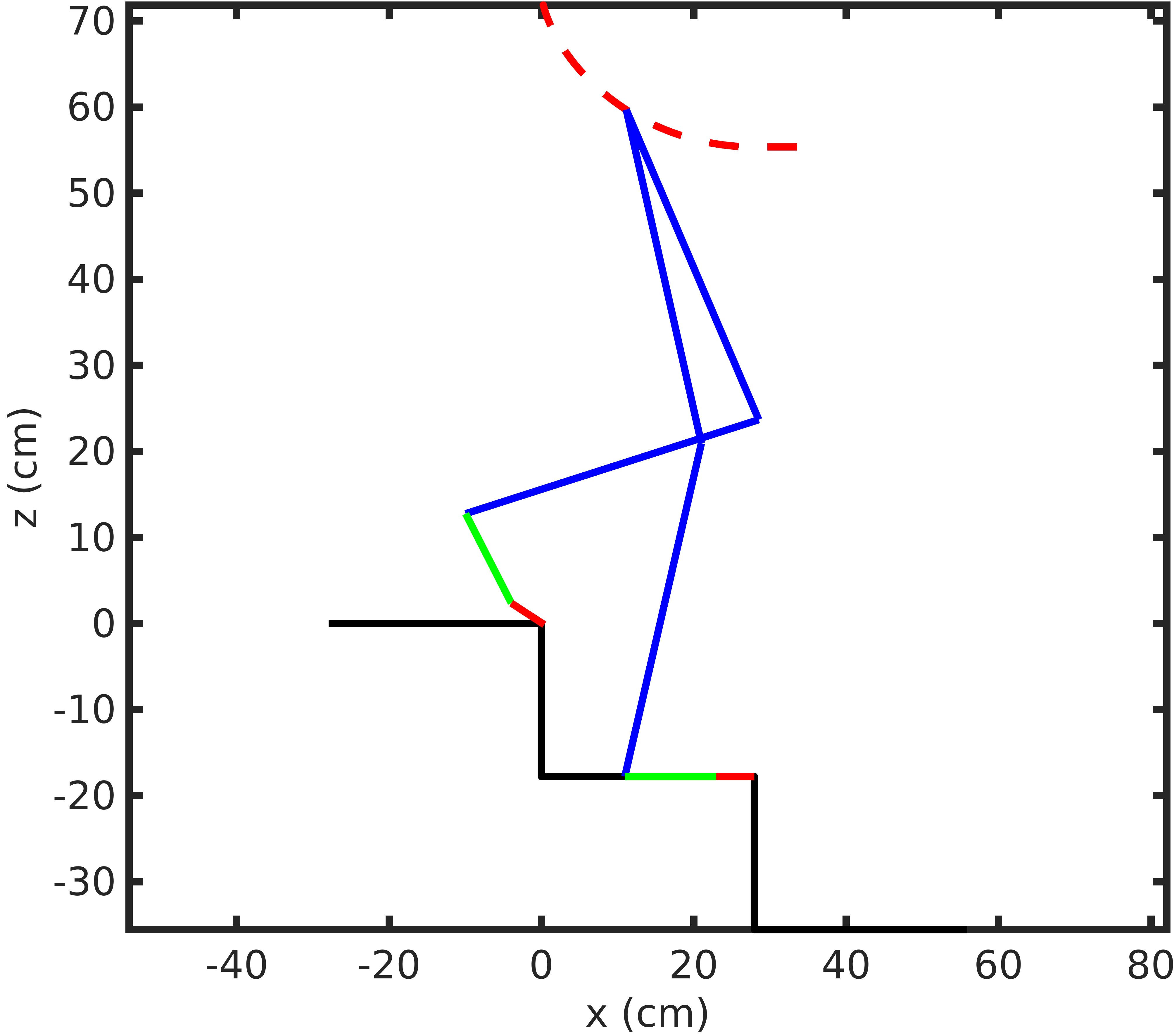}
\end{subfigure}
\begin{subfigure}{.16\textwidth}
  \centering
  \includegraphics[width=\linewidth]{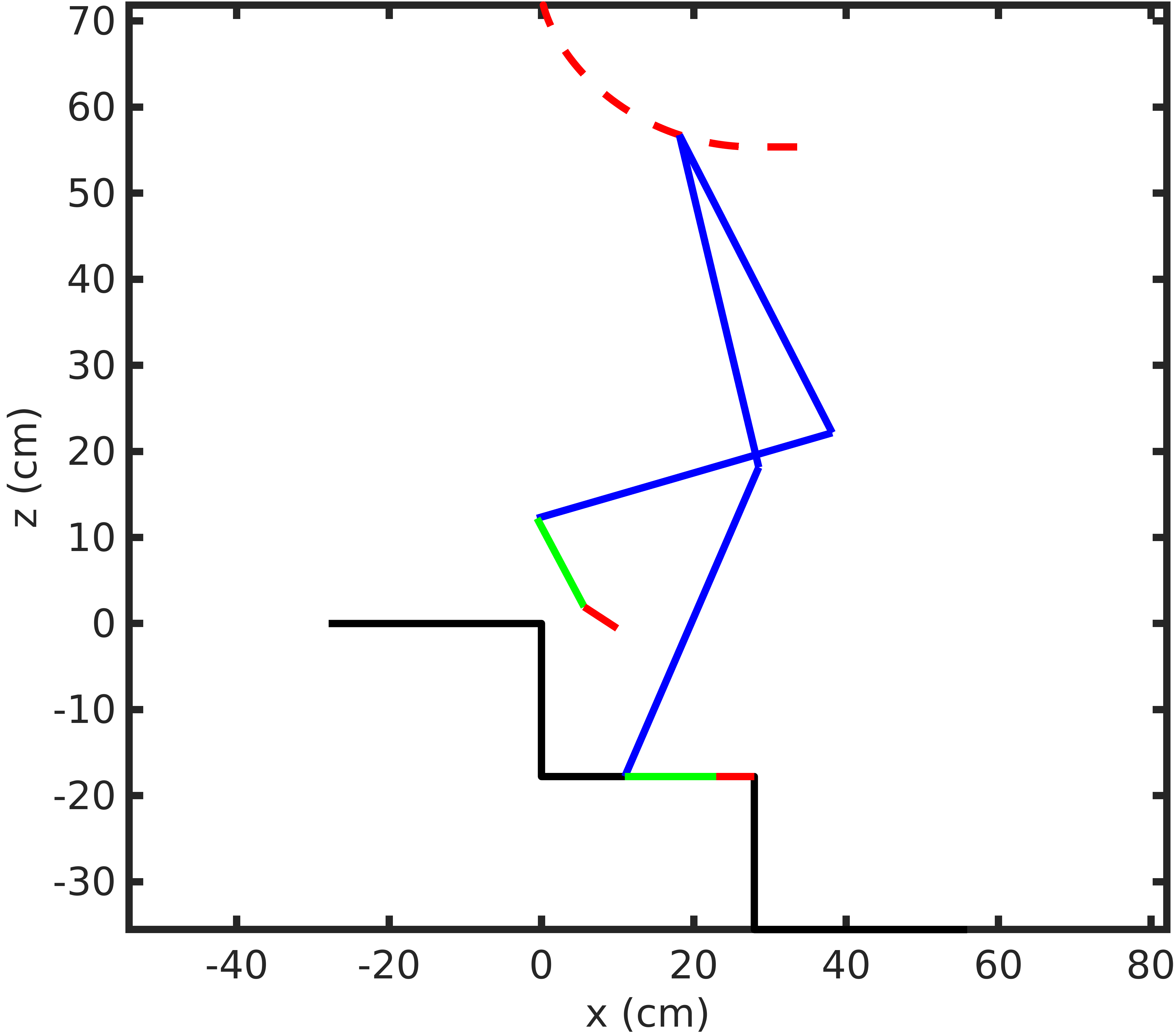}
\end{subfigure}
\begin{subfigure}{.16\textwidth}
  \centering
  \includegraphics[width=\linewidth]{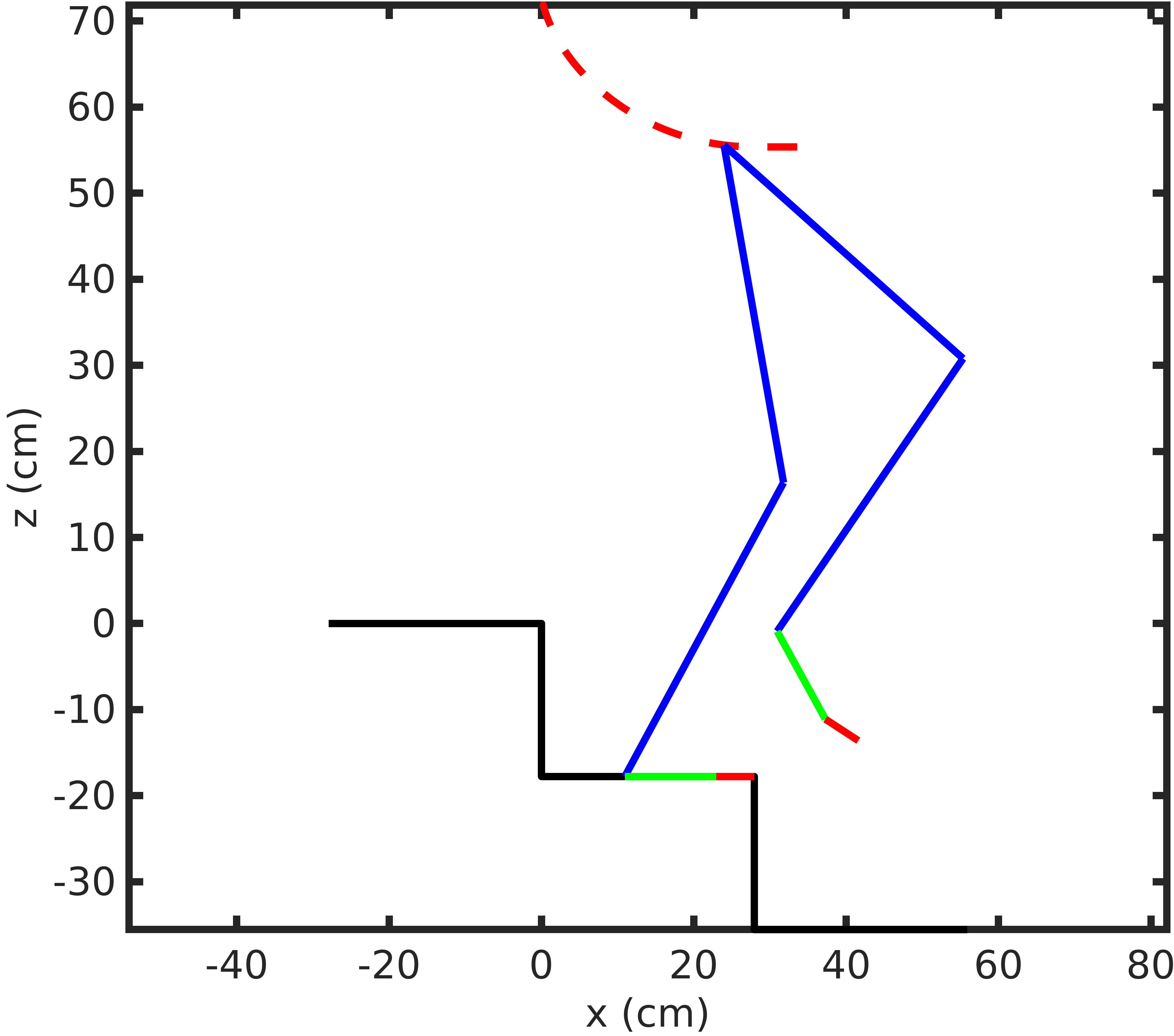}
\end{subfigure}
\begin{subfigure}{.16\textwidth}
  \centering
  \includegraphics[width=\linewidth]{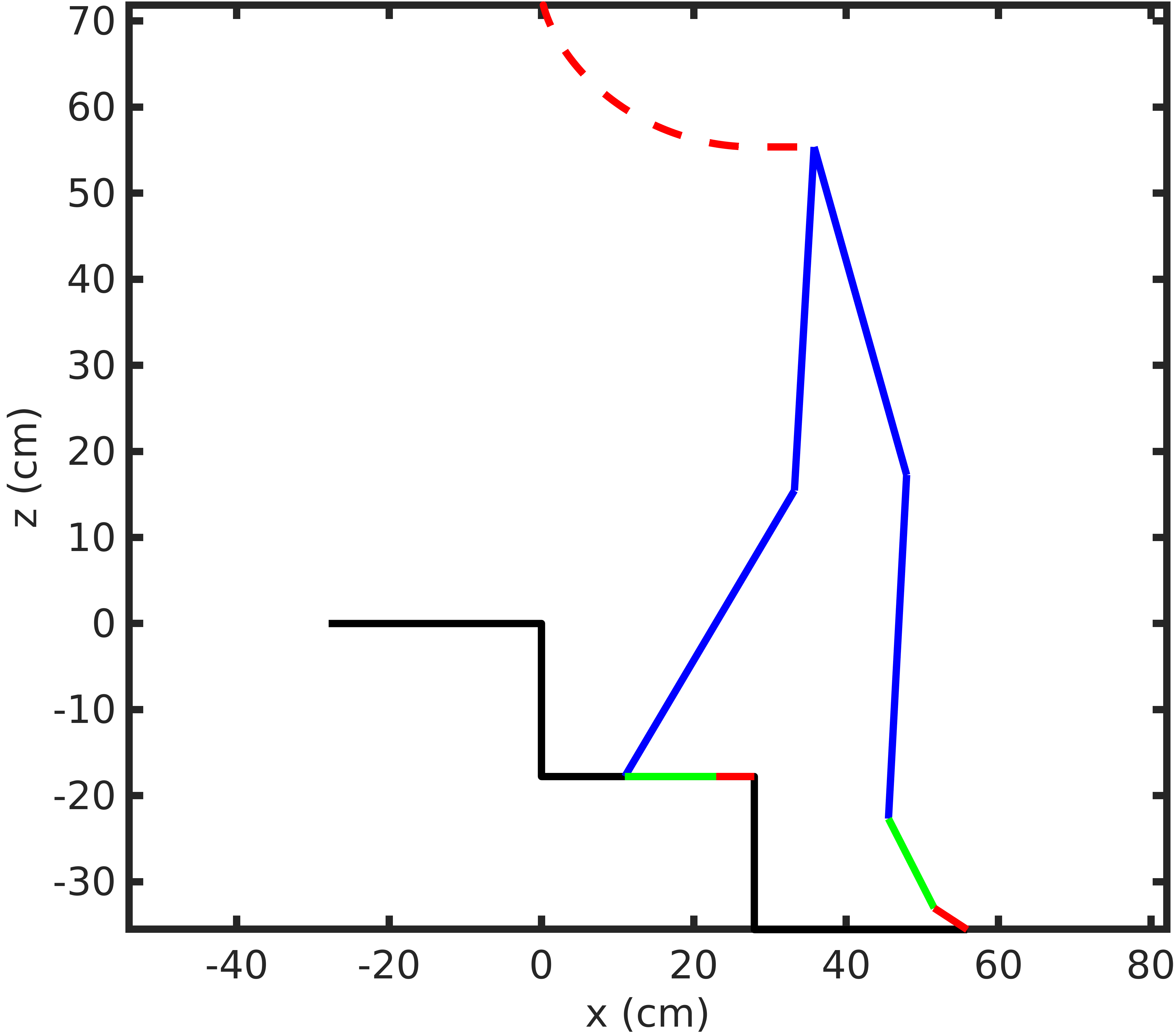}
\end{subfigure}
\begin{subfigure}{.16\textwidth}
  \centering
  \includegraphics[width=\linewidth]{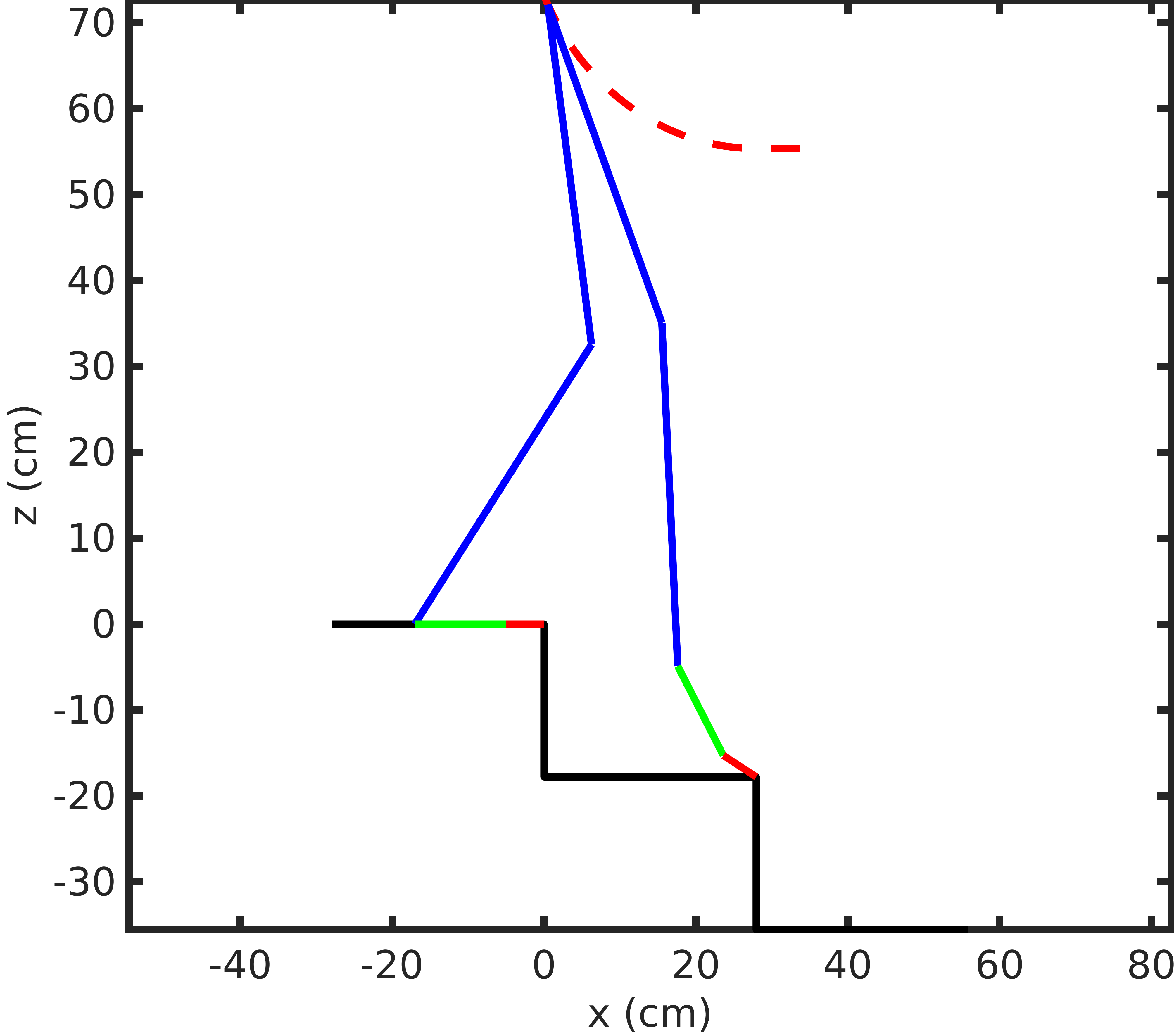}
\end{subfigure}%
\begin{subfigure}{.16\textwidth}
  \centering
  \includegraphics[width=\linewidth]{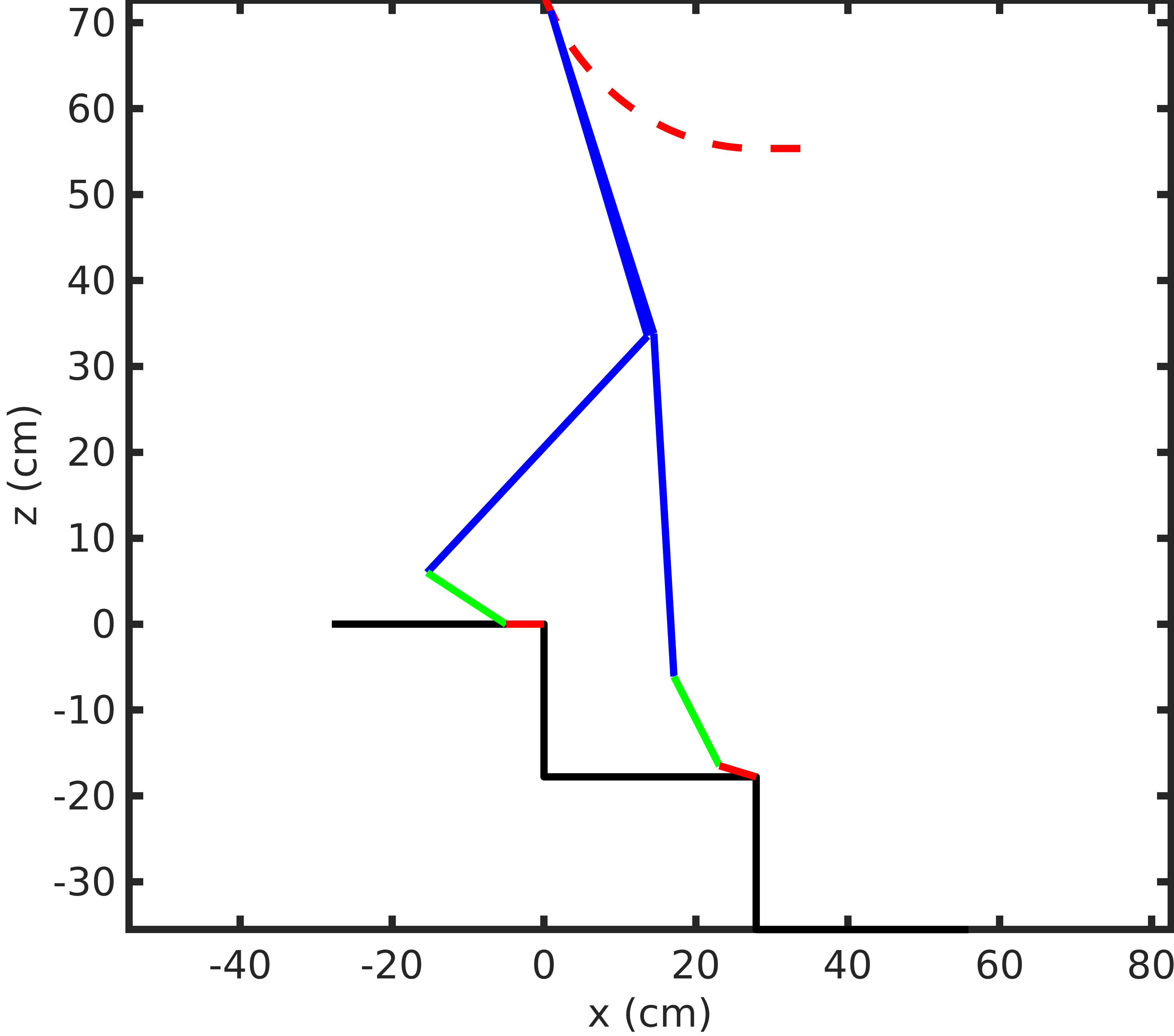}
\end{subfigure}
\begin{subfigure}{.16\textwidth}
  \centering
  \includegraphics[width=\linewidth]{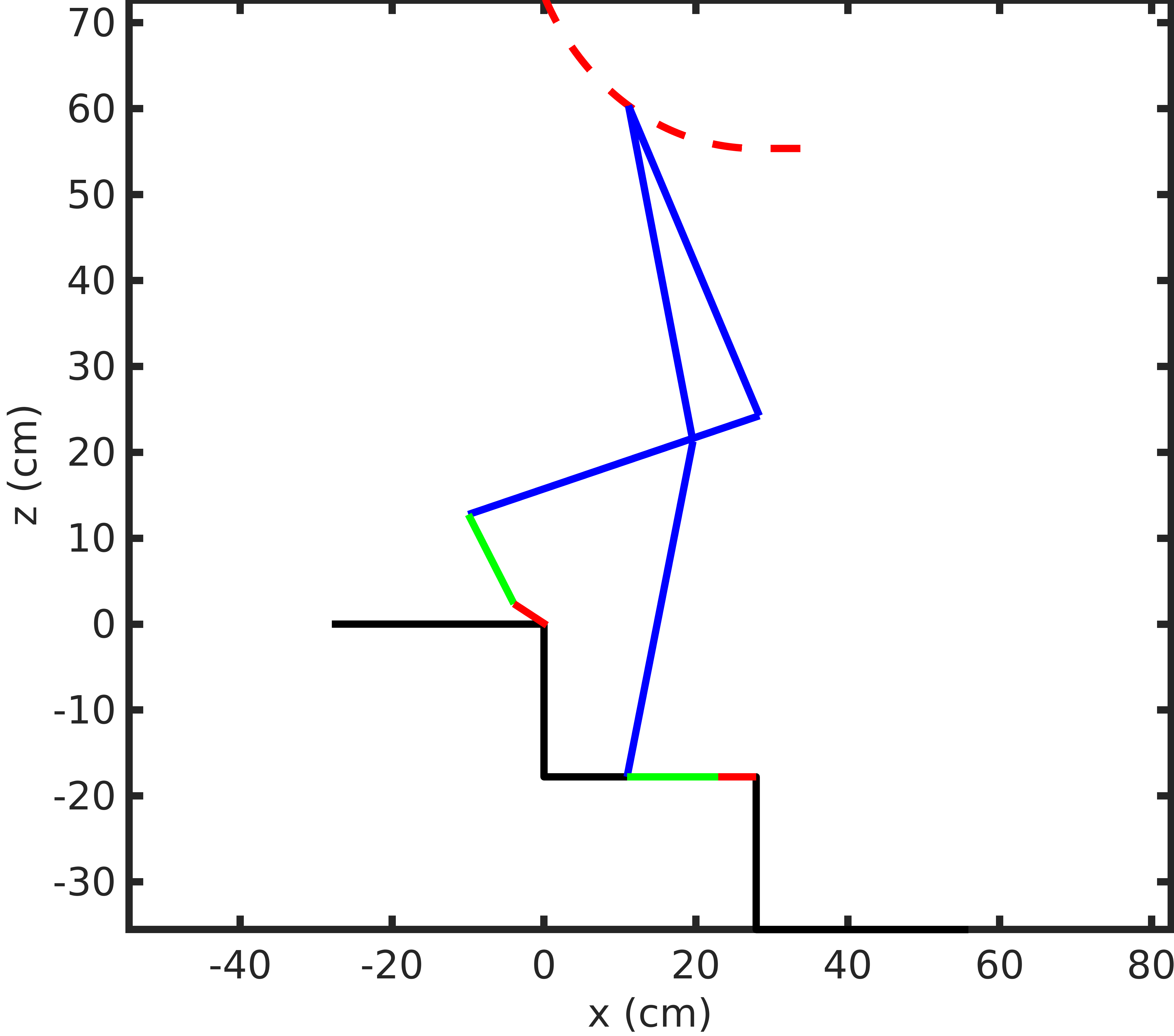}
\end{subfigure}
\begin{subfigure}{.16\textwidth}
  \centering
  \includegraphics[width=\linewidth]{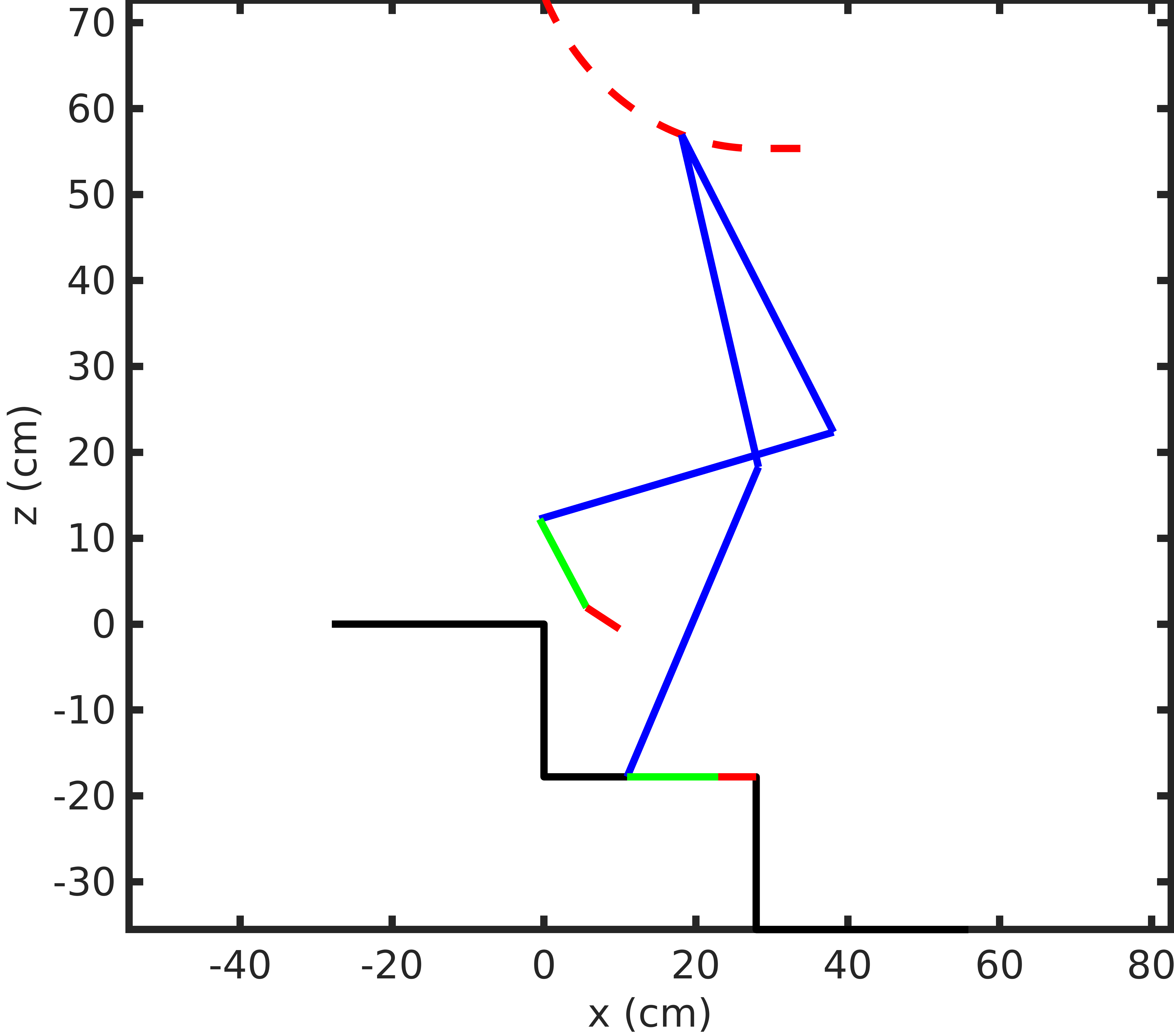}
\end{subfigure}
\begin{subfigure}{.16\textwidth}
  \centering
  \includegraphics[width=\linewidth]{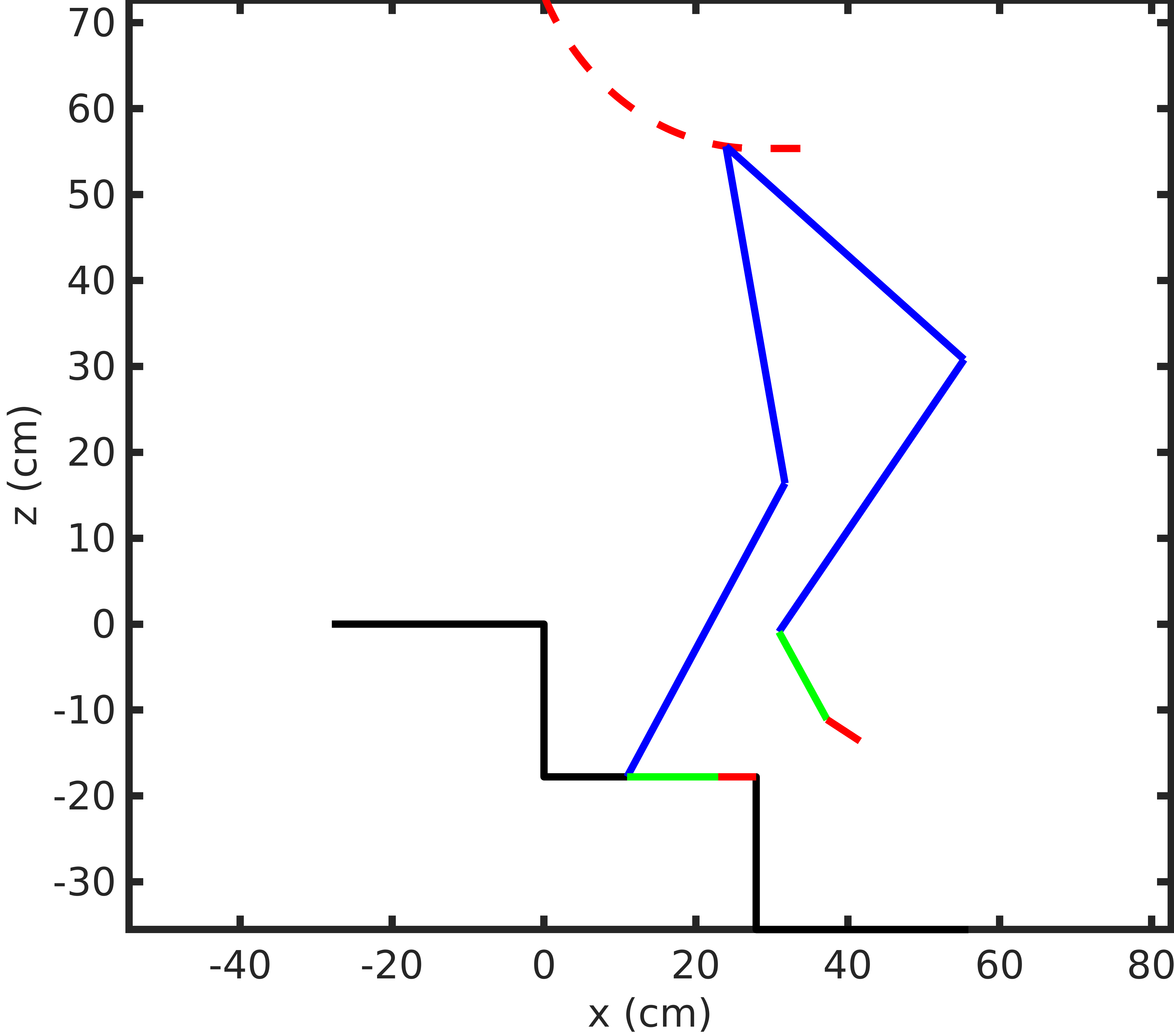}
\end{subfigure}
\begin{subfigure}{.16\textwidth}
  \centering
  \includegraphics[width=\linewidth]{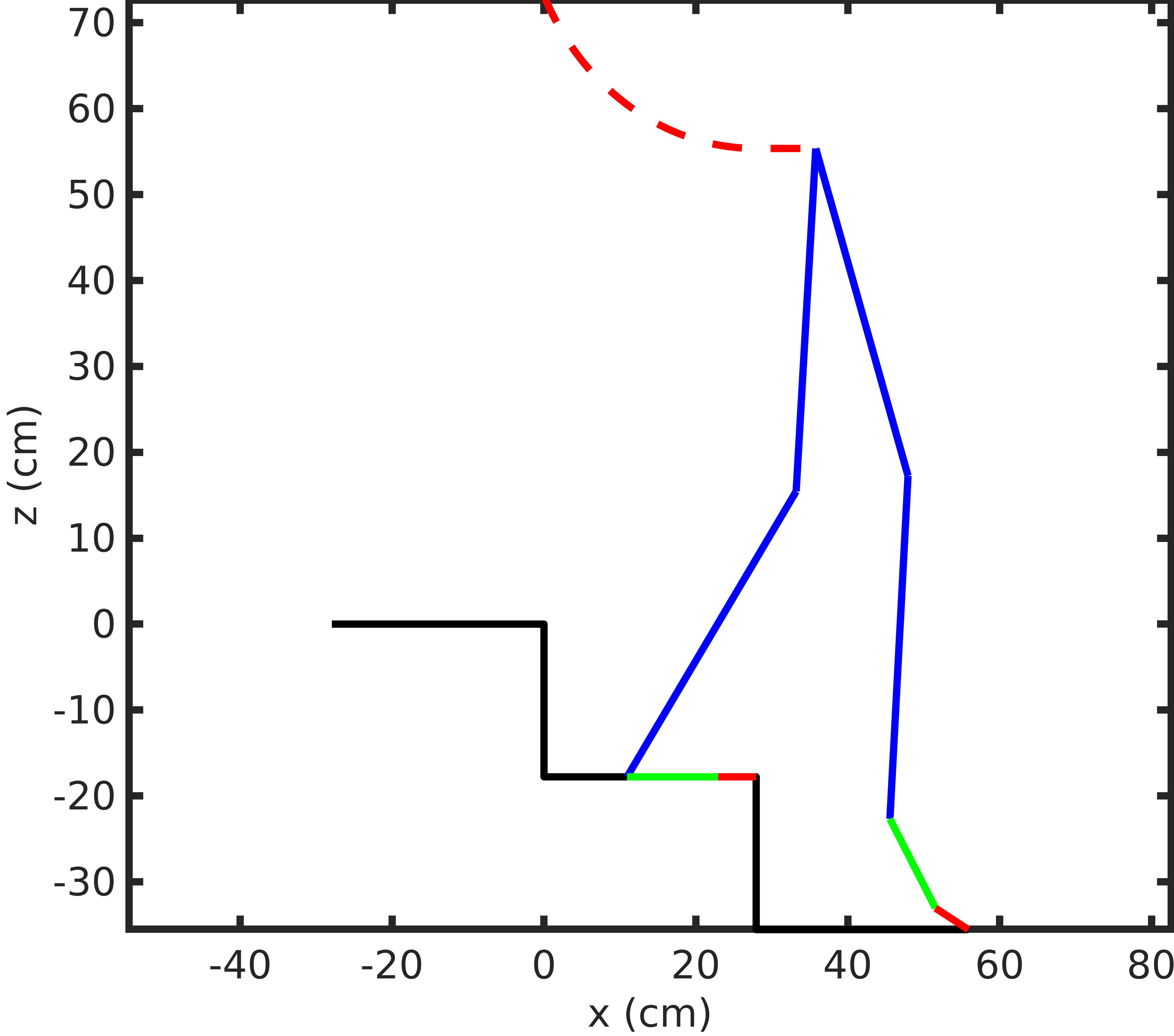}
\end{subfigure}
\begin{subfigure}{.16\textwidth}
  \centering
  \includegraphics[width=\linewidth]{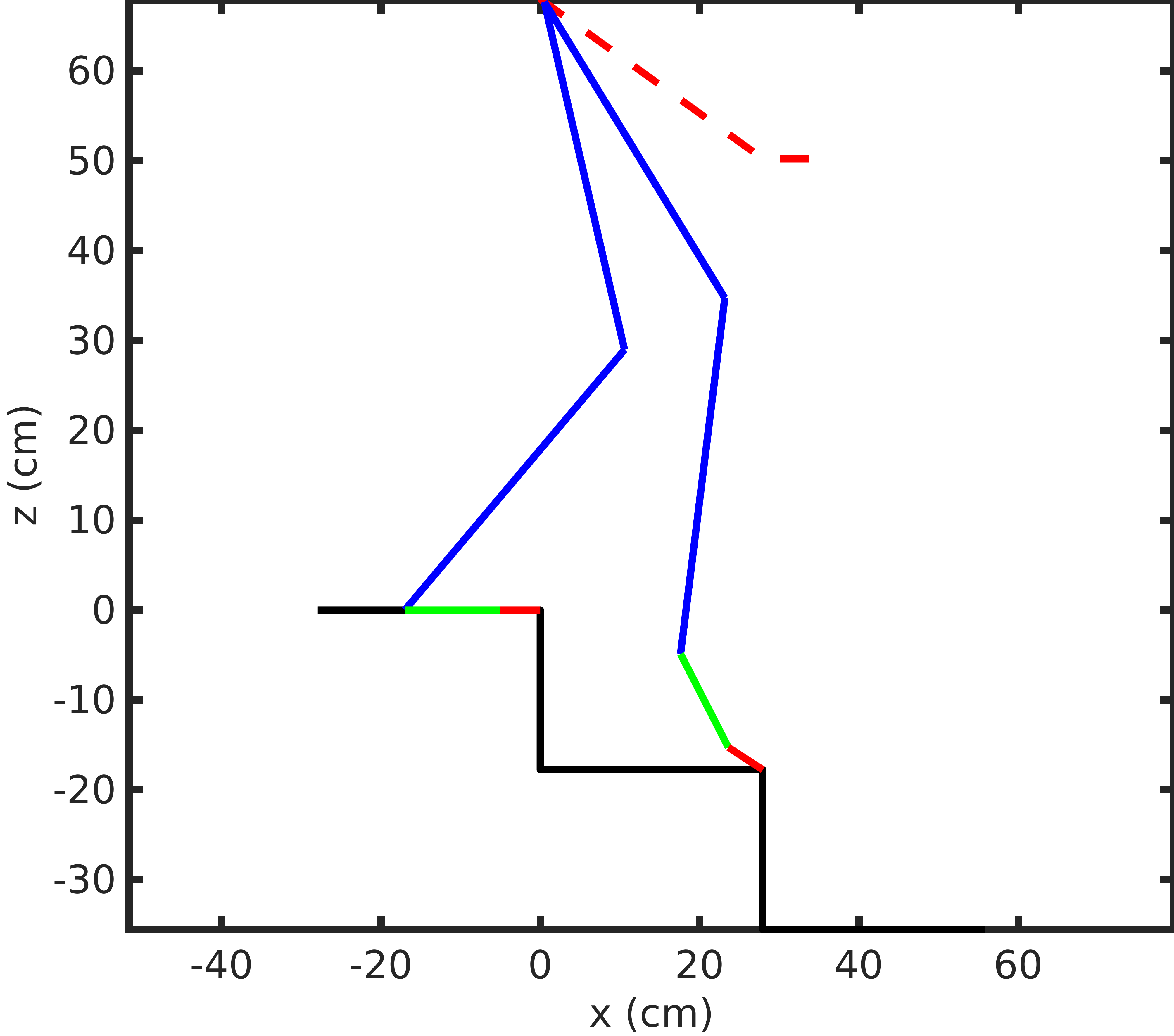}
\end{subfigure}%
\begin{subfigure}{.16\textwidth}
  \centering
  \includegraphics[width=\linewidth]{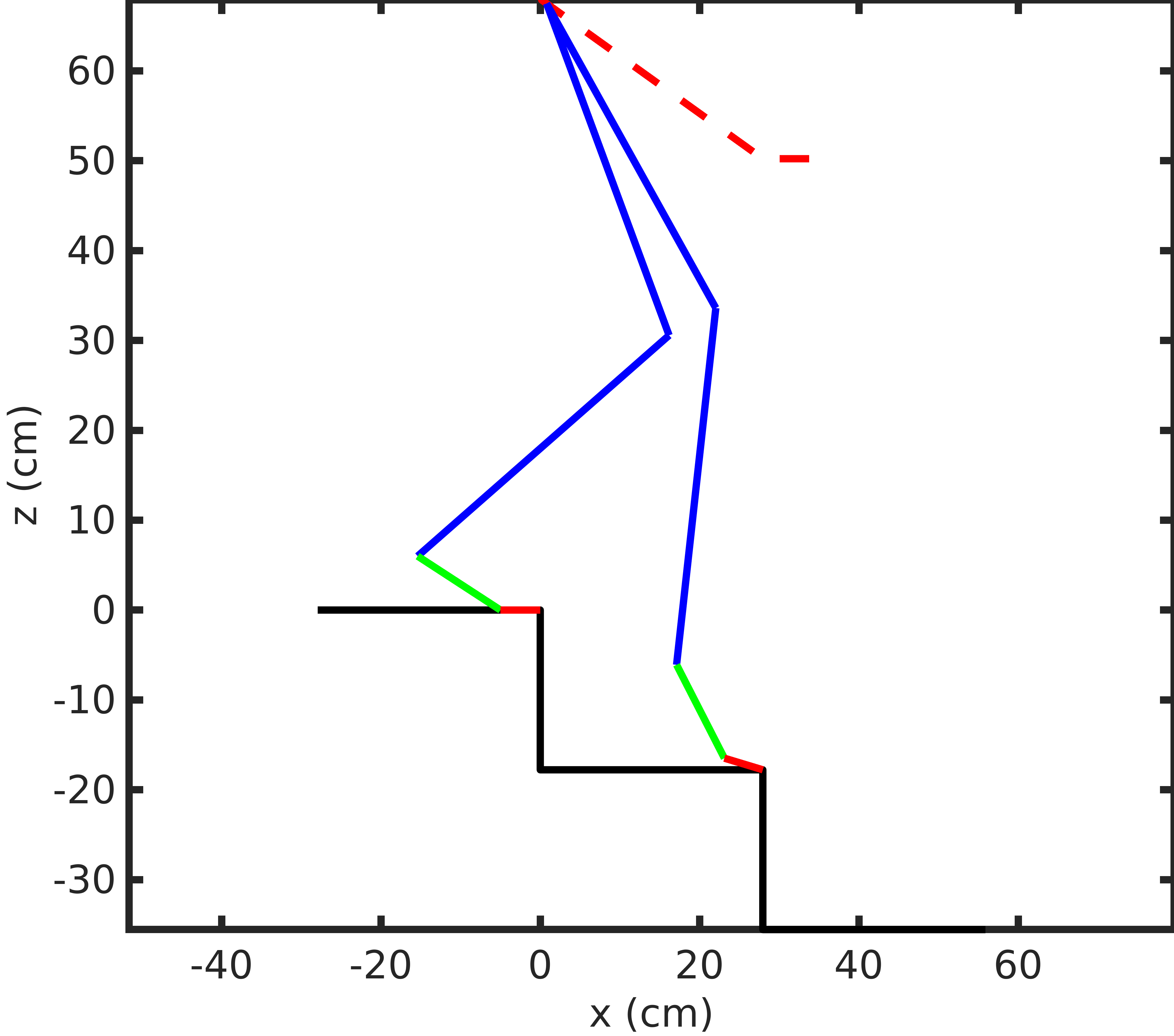}
\end{subfigure}
\begin{subfigure}{.16\textwidth}
  \centering
  \includegraphics[width=\linewidth]{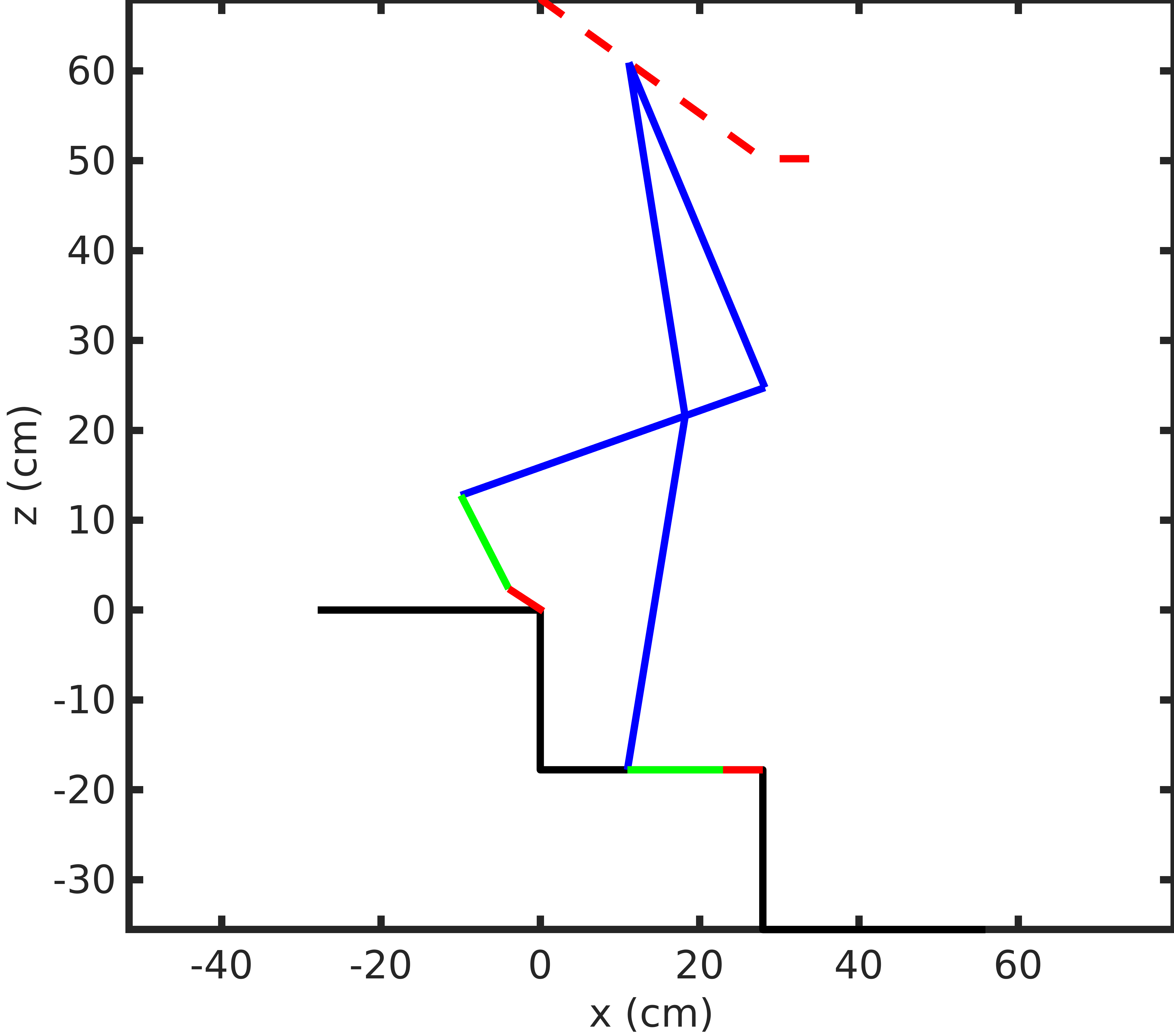}
\end{subfigure}
\begin{subfigure}{.16\textwidth}
  \centering
  \includegraphics[width=\linewidth]{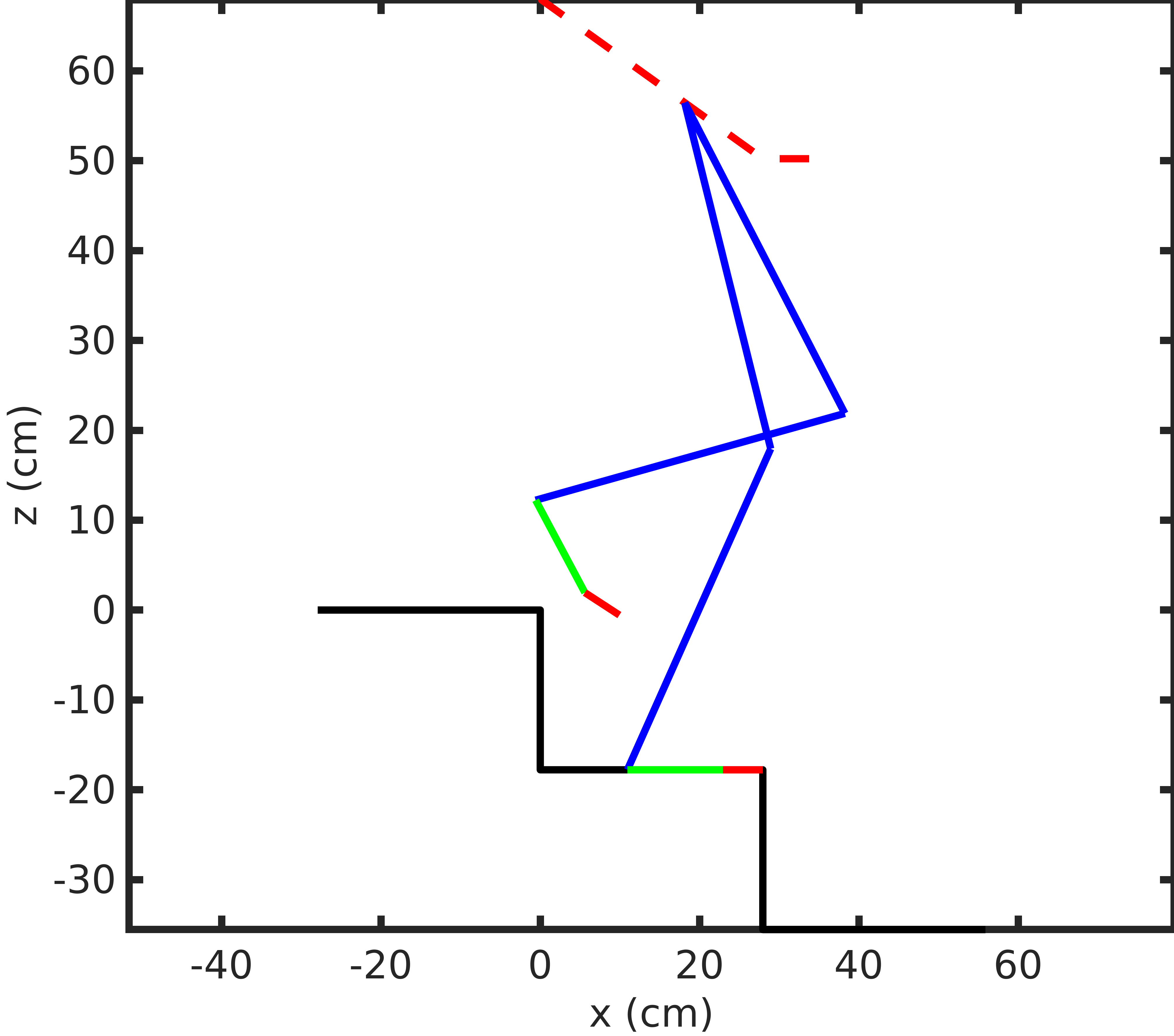}
\end{subfigure}
\begin{subfigure}{.16\textwidth}
  \centering
  \includegraphics[width=\linewidth]{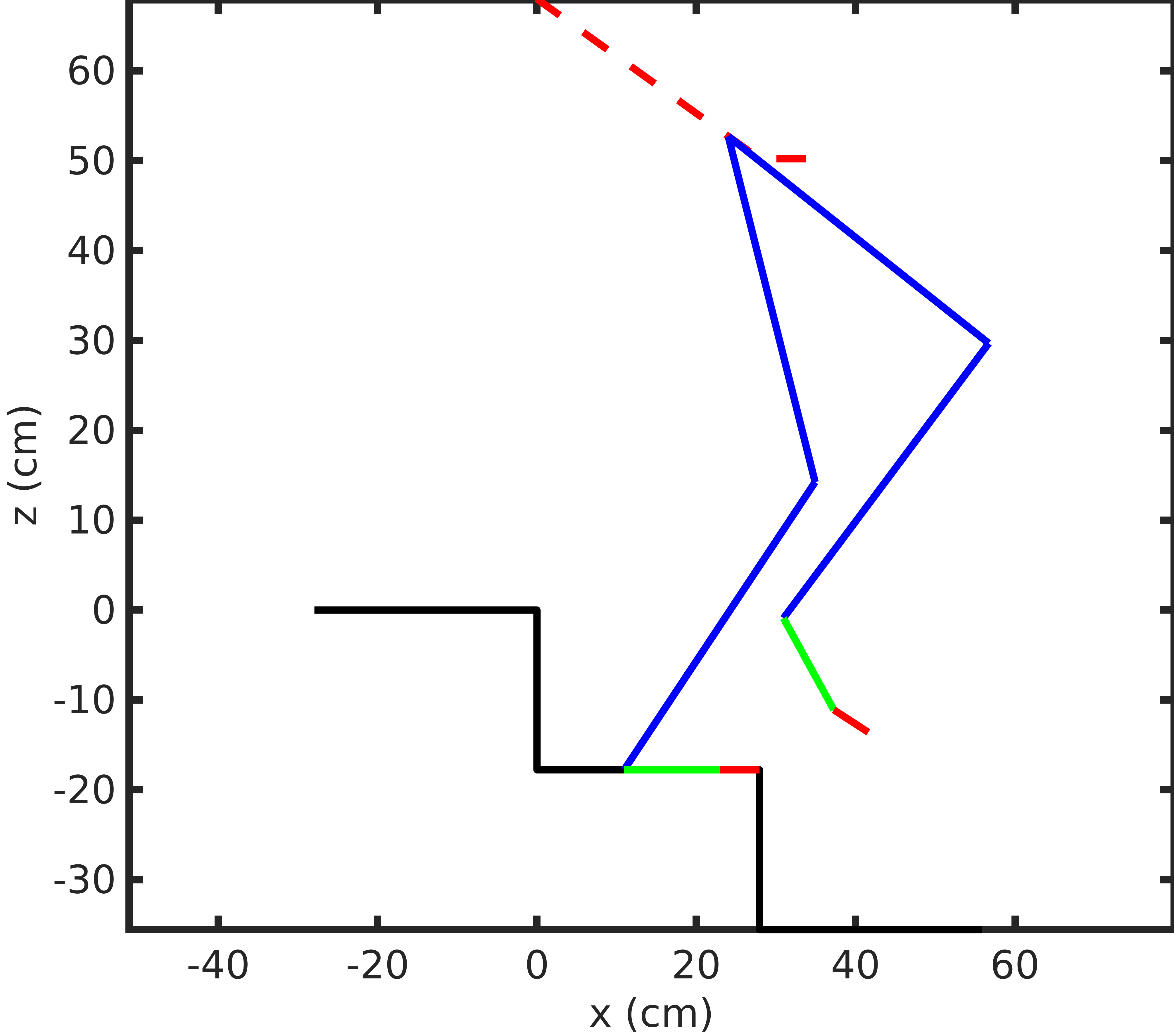}
\end{subfigure}
\begin{subfigure}{.16\textwidth}
  \centering
  \includegraphics[width=\linewidth]{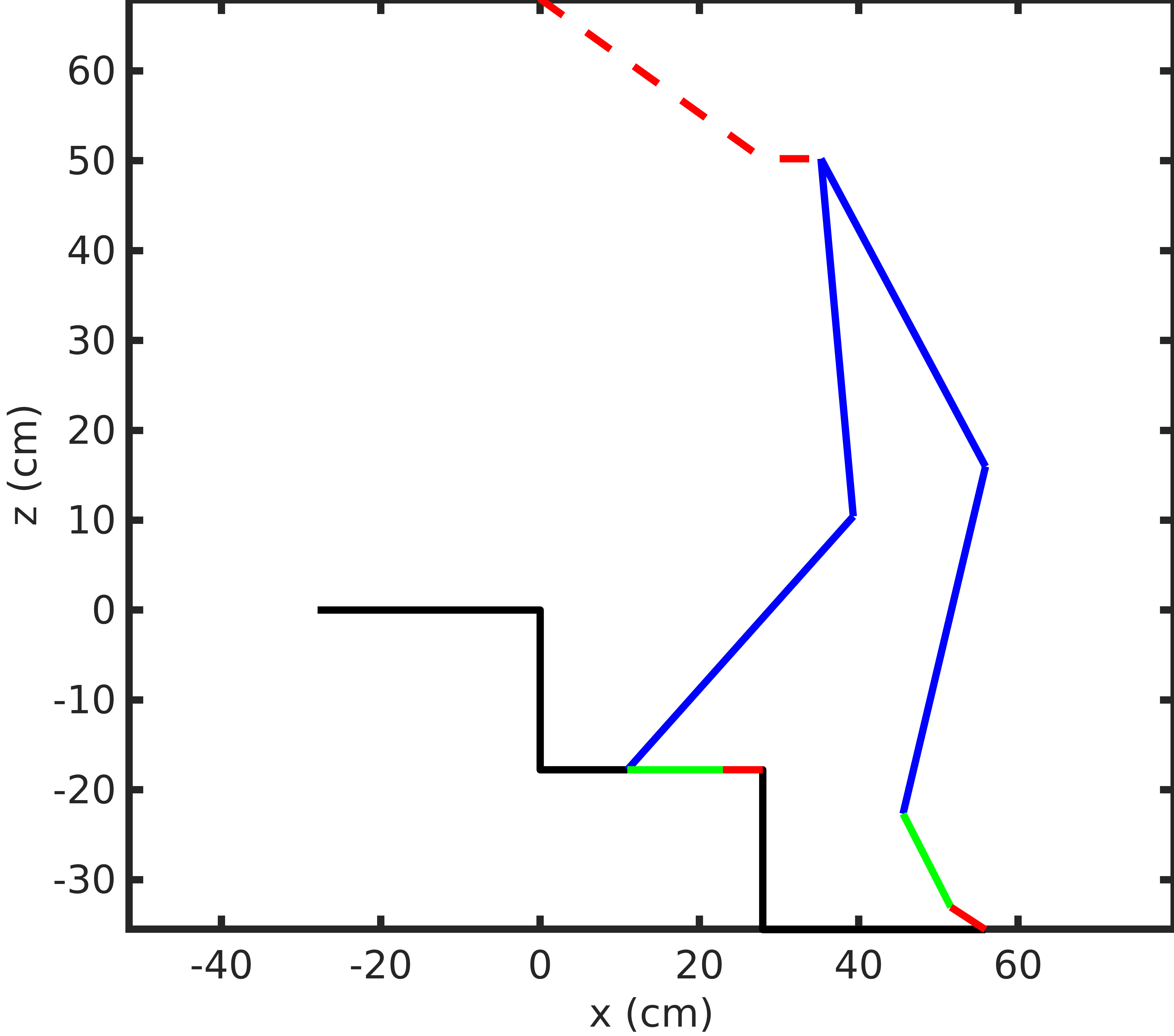}
\end{subfigure}
\caption{Motion Plots of Intermediate Instants. The first row corresponds to the brachistochrone hip trajectory, the second row for the circular arc trajectory and the thros row for the virtual slope method}
\label{fig:motion}
\end{figure*}

\begin{figure*}
\begin{subfigure}{.242\textwidth}
  \centering
  \includegraphics[width=0.95\linewidth]{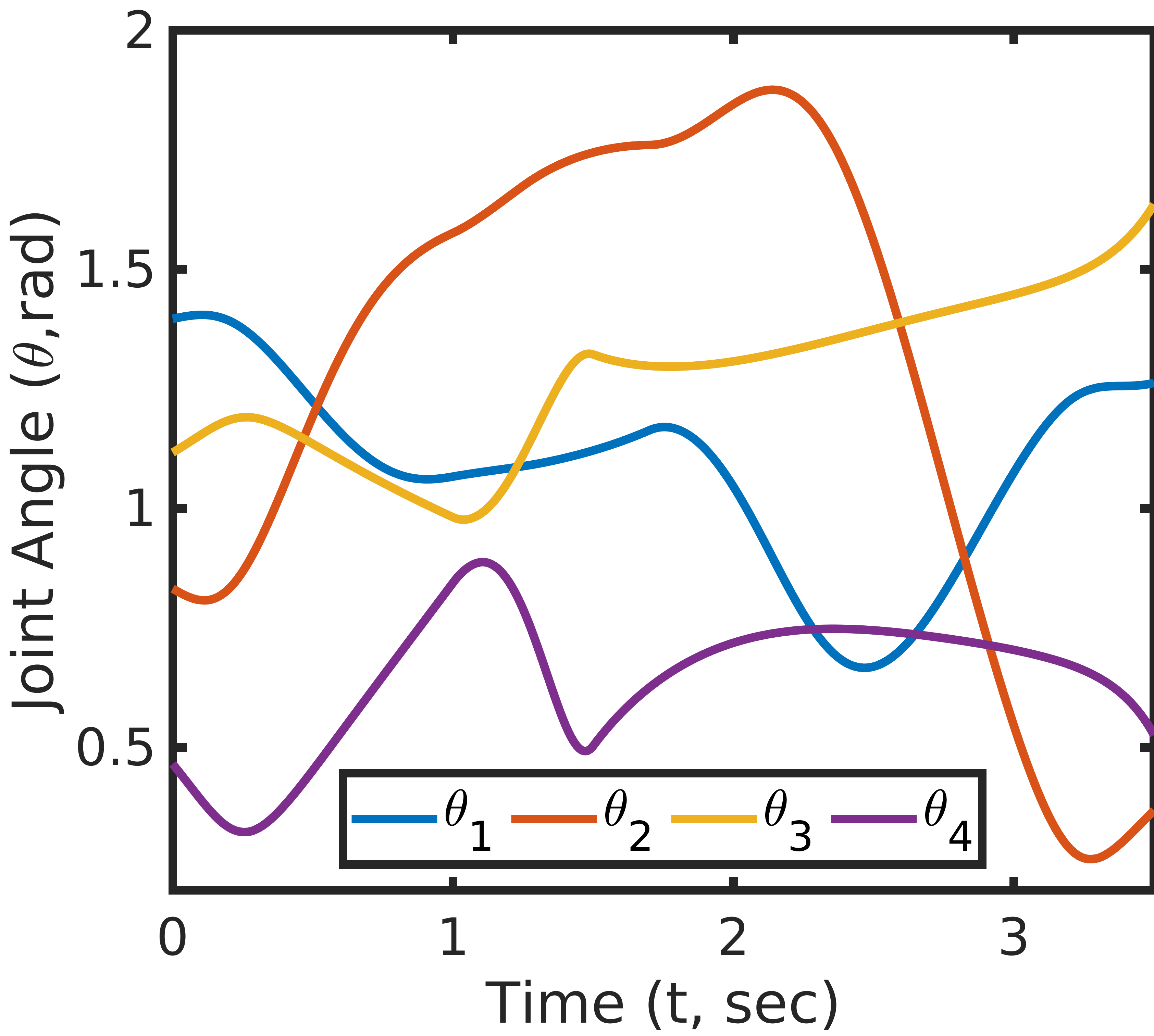}
\end{subfigure}%
\begin{subfigure}{.242\textwidth}
  \centering
  \includegraphics[width=0.95\linewidth]{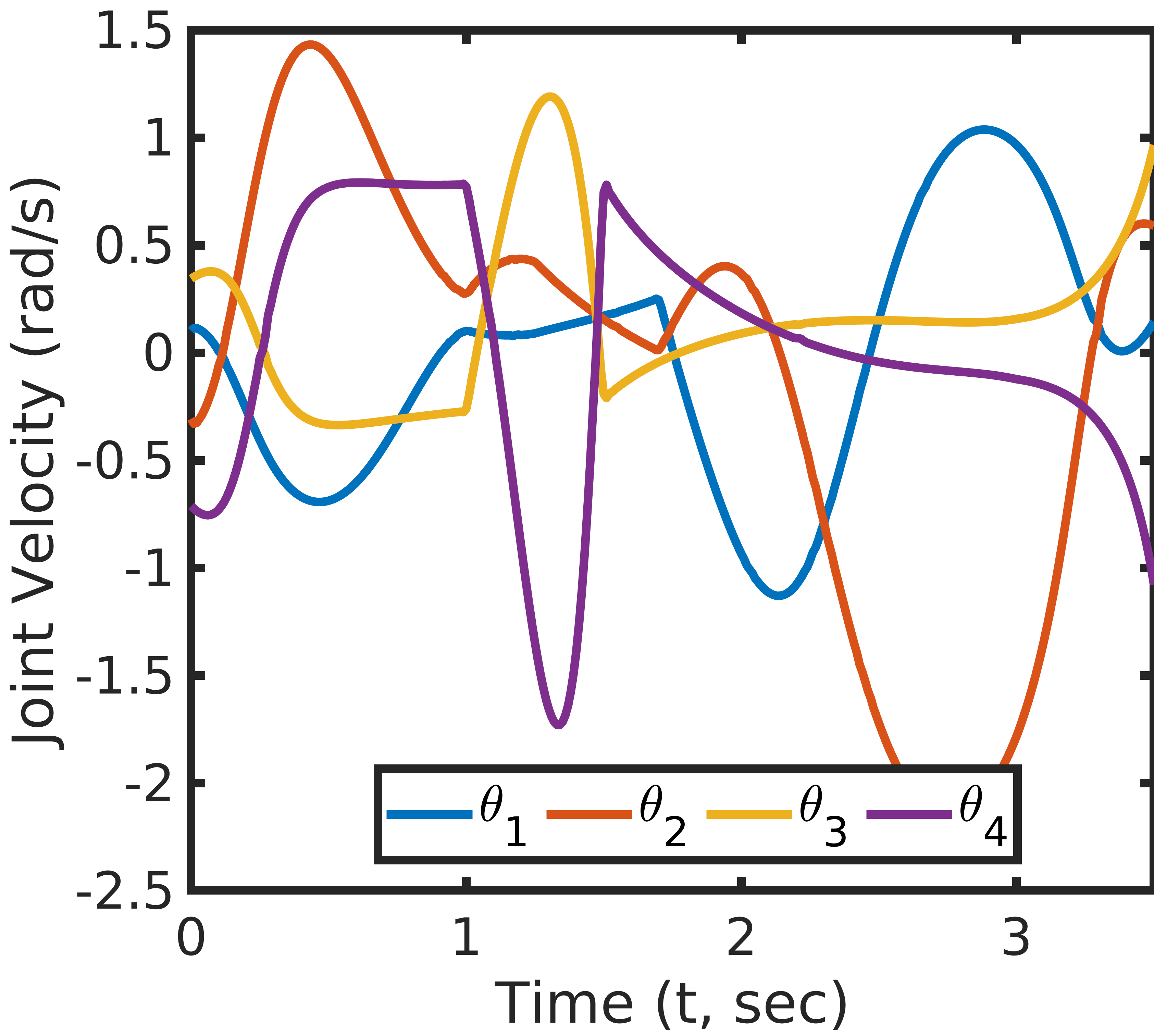}
\end{subfigure}
\begin{subfigure}{.242\textwidth}
  \centering
  \includegraphics[width=0.95\linewidth]{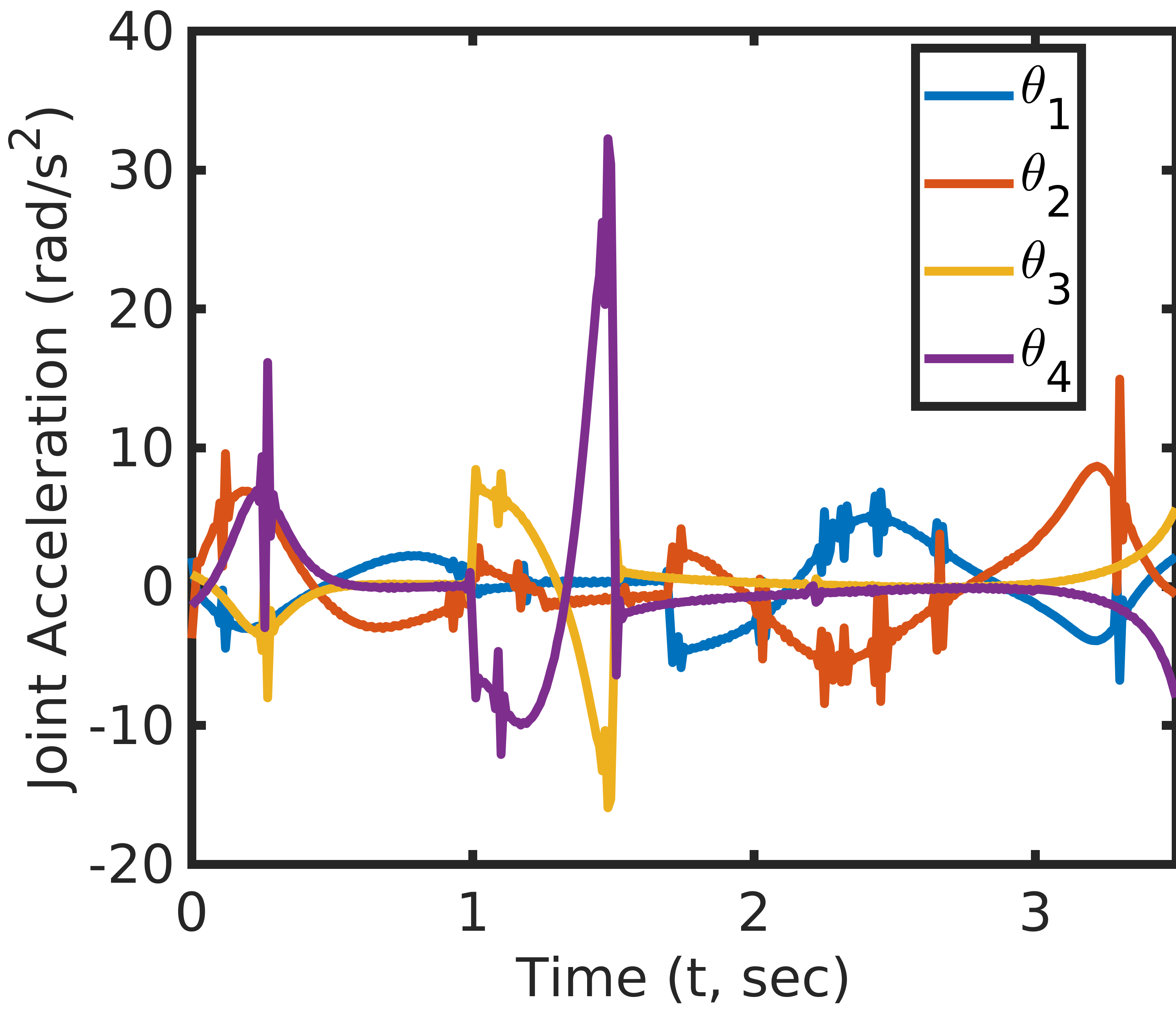}
\end{subfigure}
\begin{subfigure}{.242\textwidth}
  \centering
  \includegraphics[width=0.95\linewidth]{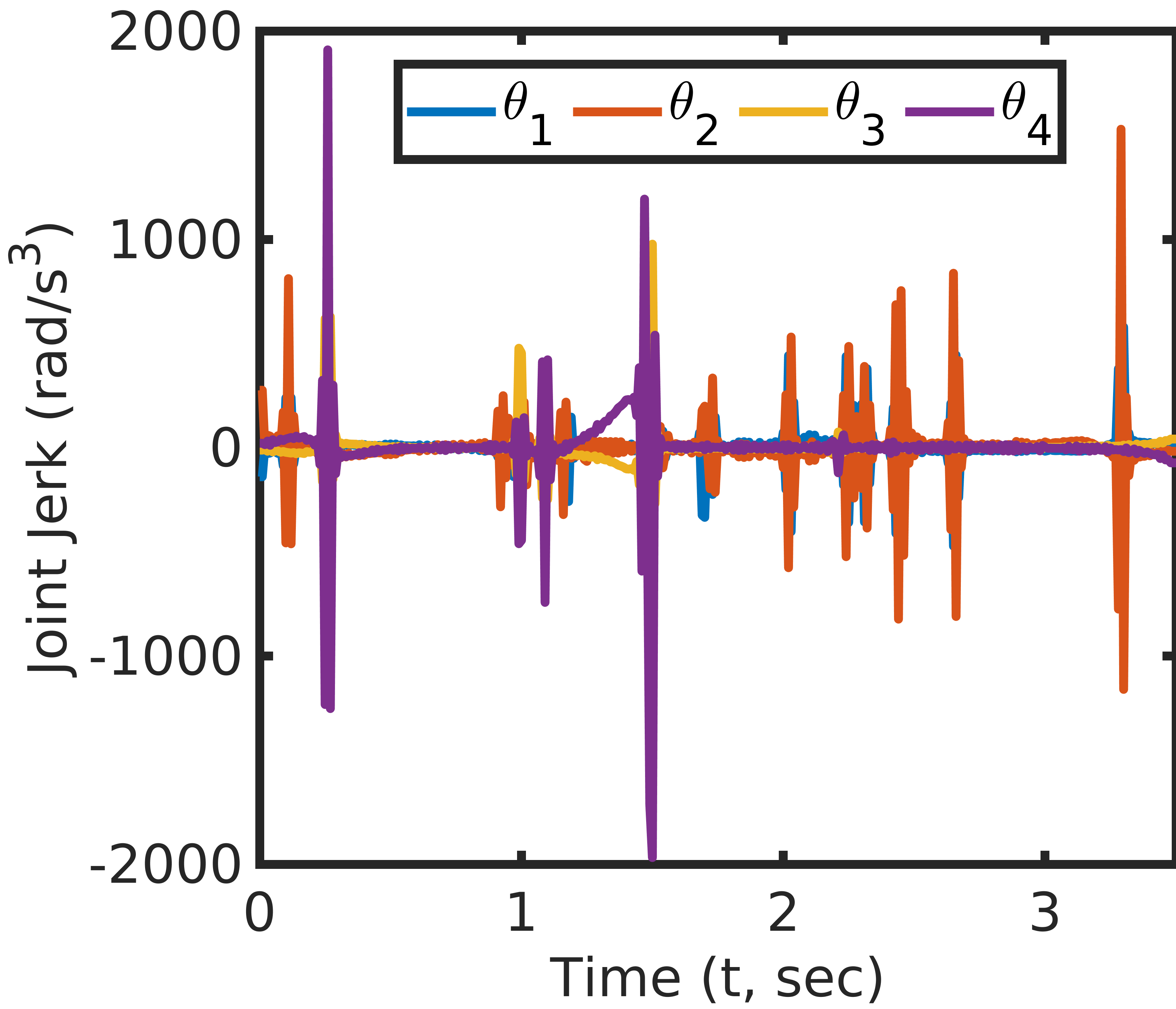}
\end{subfigure}
\begin{subfigure}{.242\textwidth}
  \centering
  \includegraphics[width=0.95\linewidth]{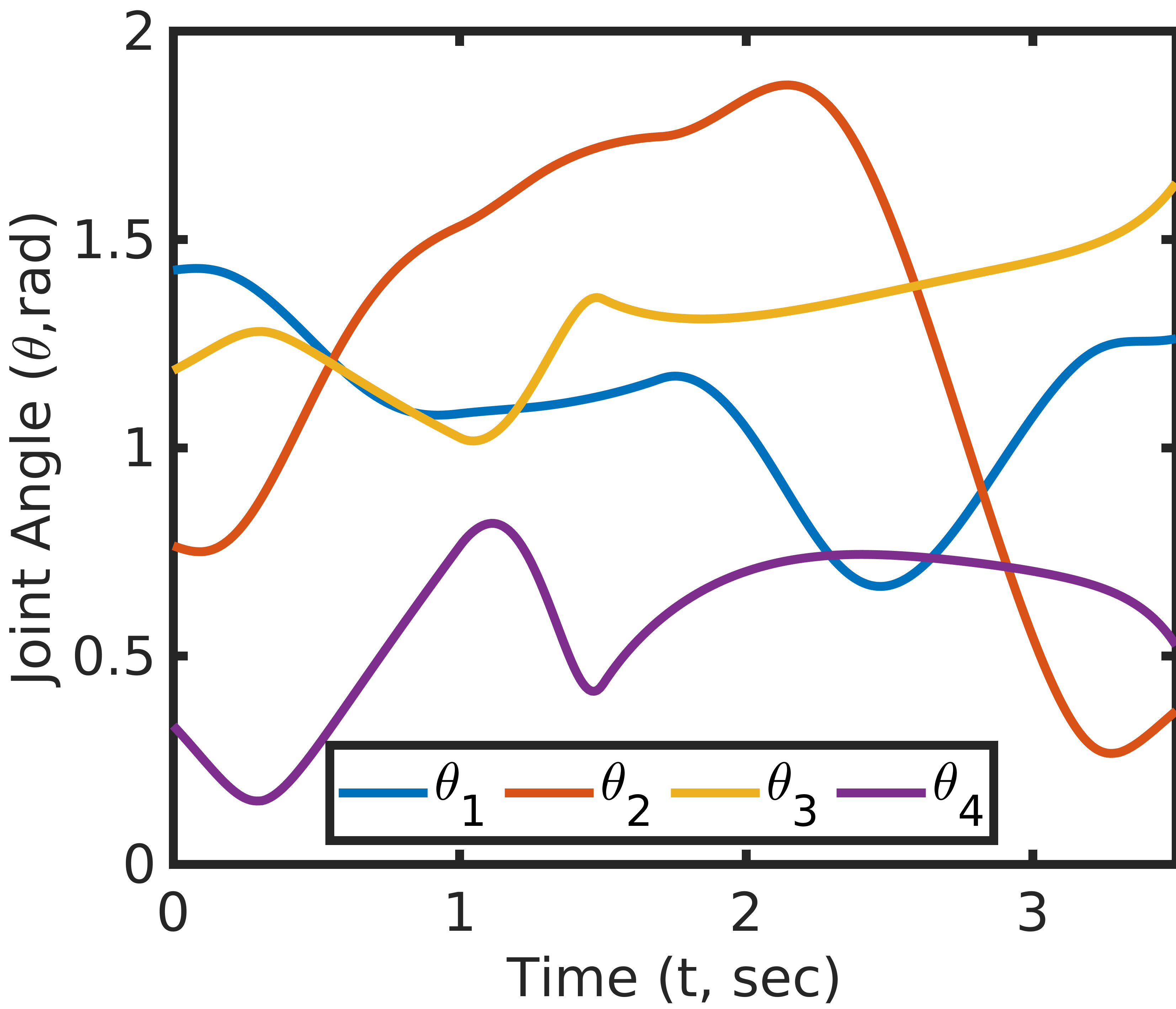}
\end{subfigure}%
\begin{subfigure}{.242\textwidth}
  \centering
  \includegraphics[width=0.95\linewidth]{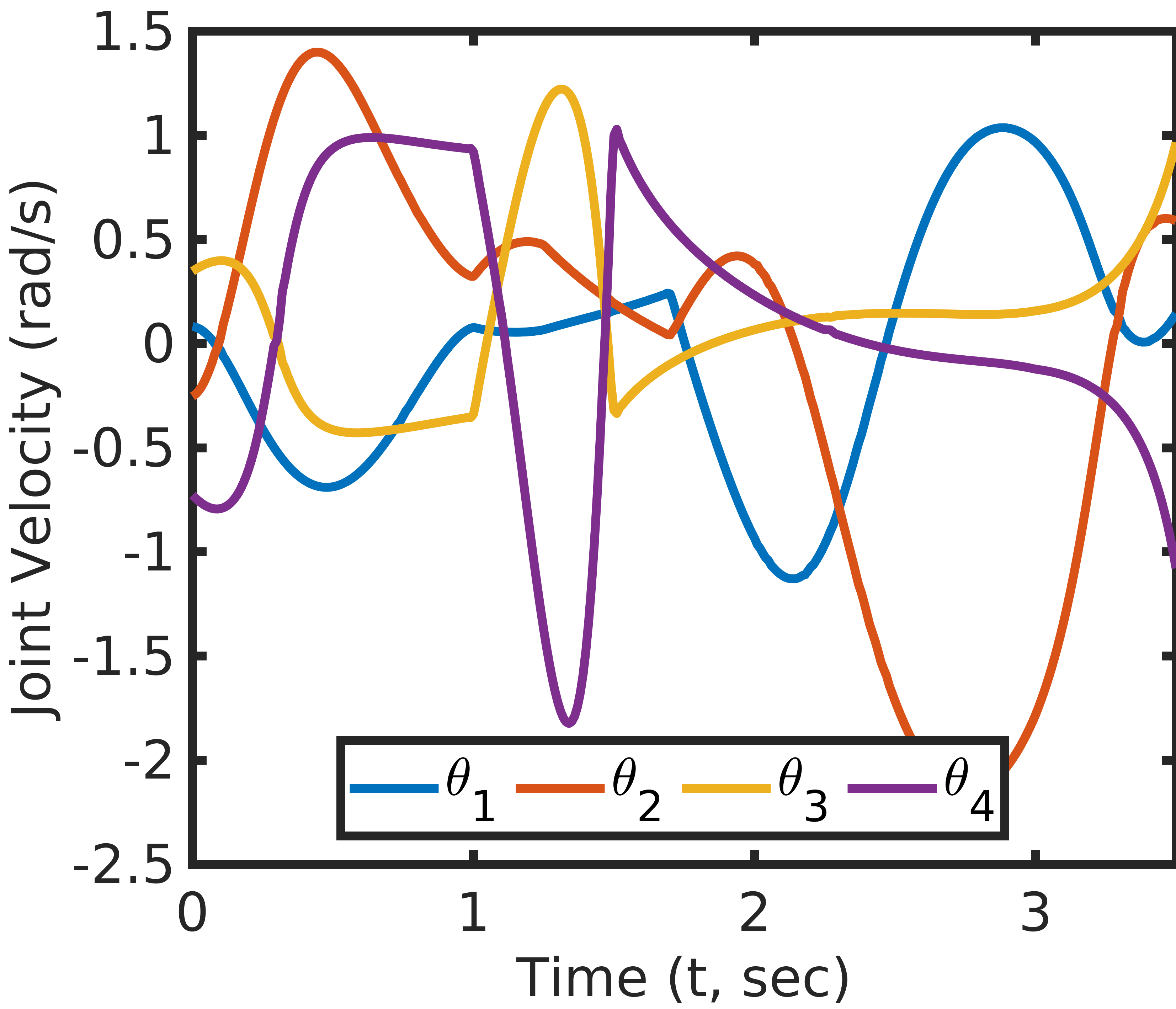}
\end{subfigure}
\begin{subfigure}{.242\textwidth}
  \centering
  \includegraphics[width=0.95\linewidth]{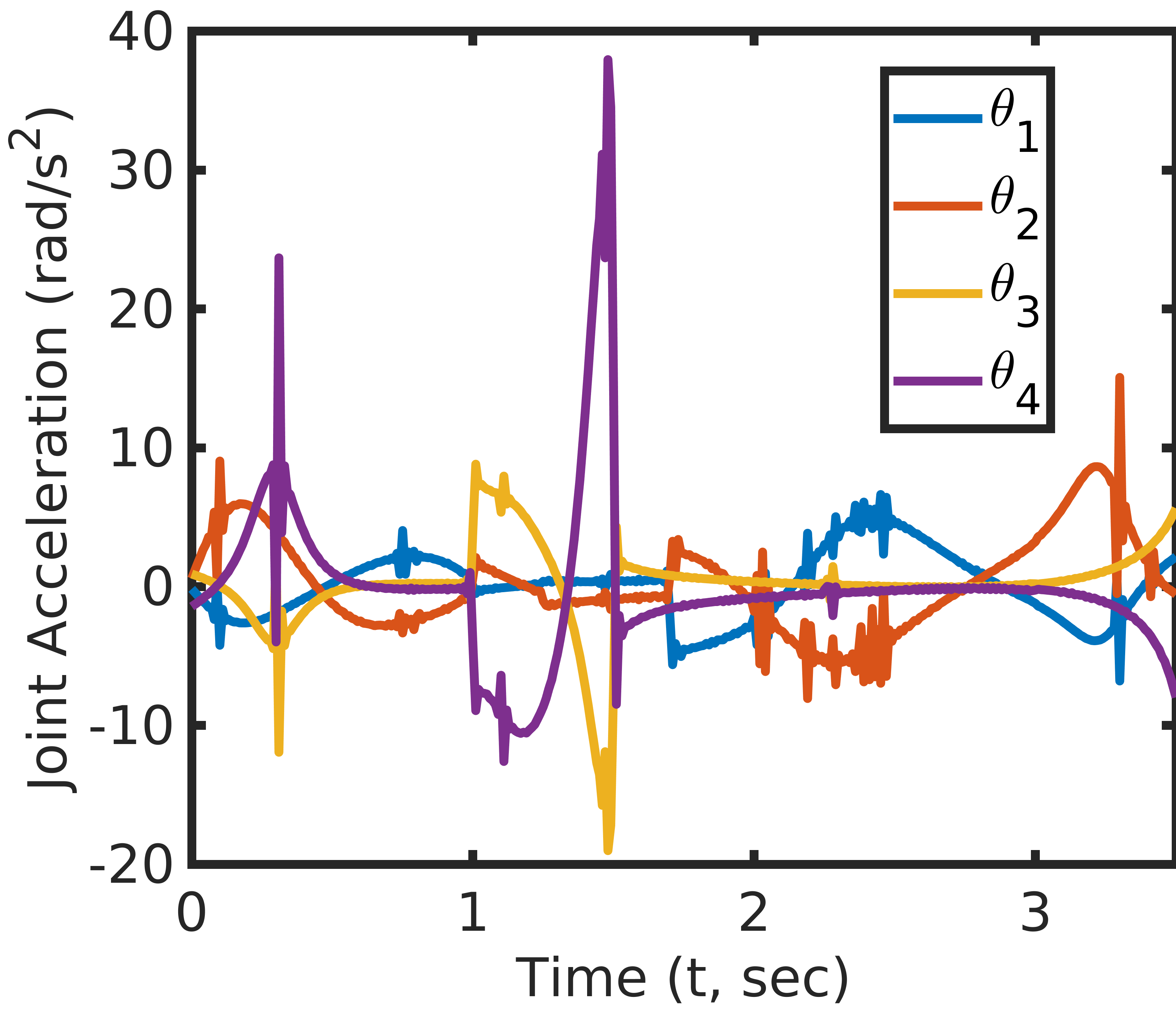}
\end{subfigure}
\begin{subfigure}{.242\textwidth}
  \centering
  \includegraphics[width=0.95\linewidth]{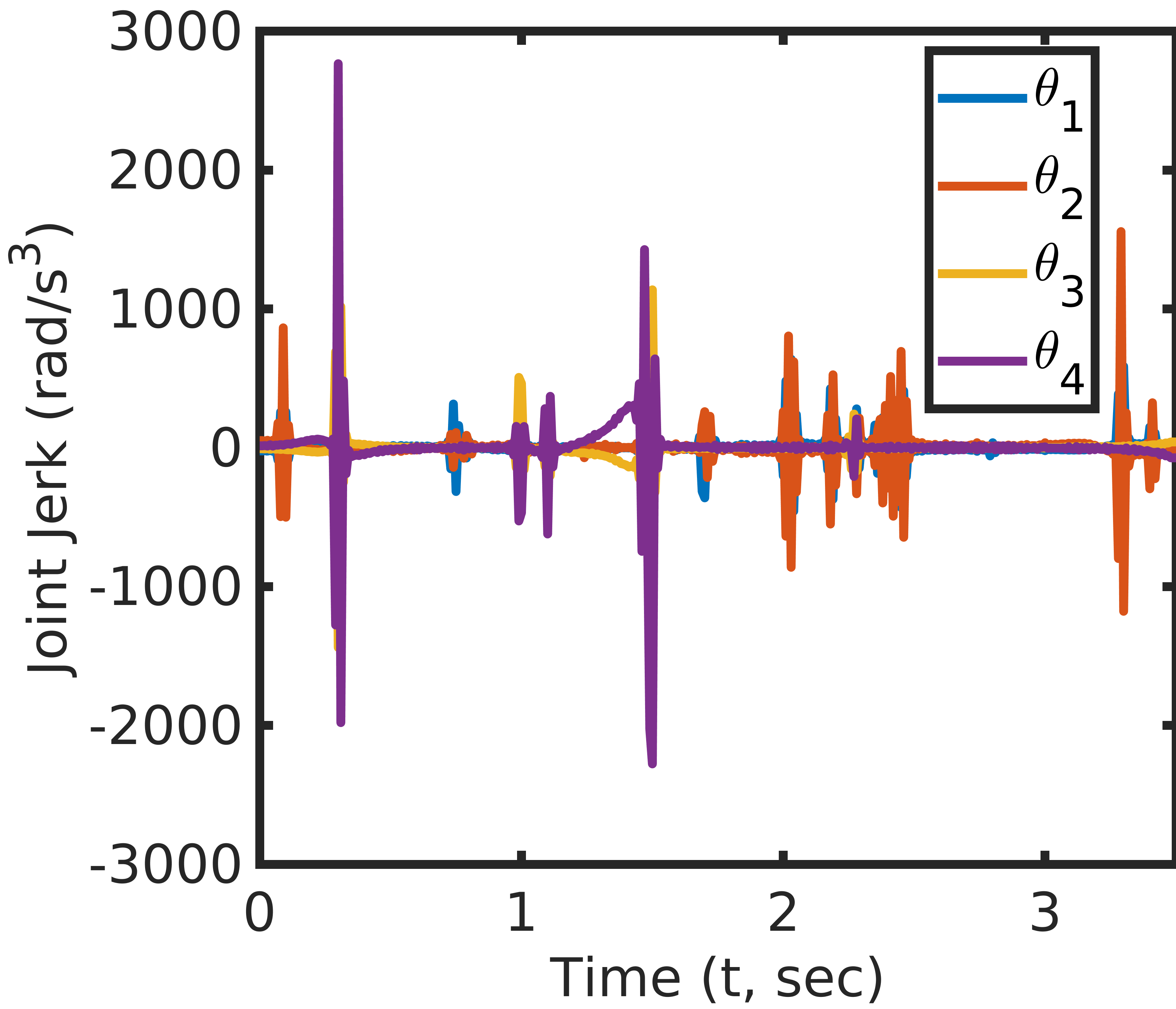}
\end{subfigure}
\begin{subfigure}{.242\textwidth}
  \centering
  \includegraphics[width=0.95\linewidth]{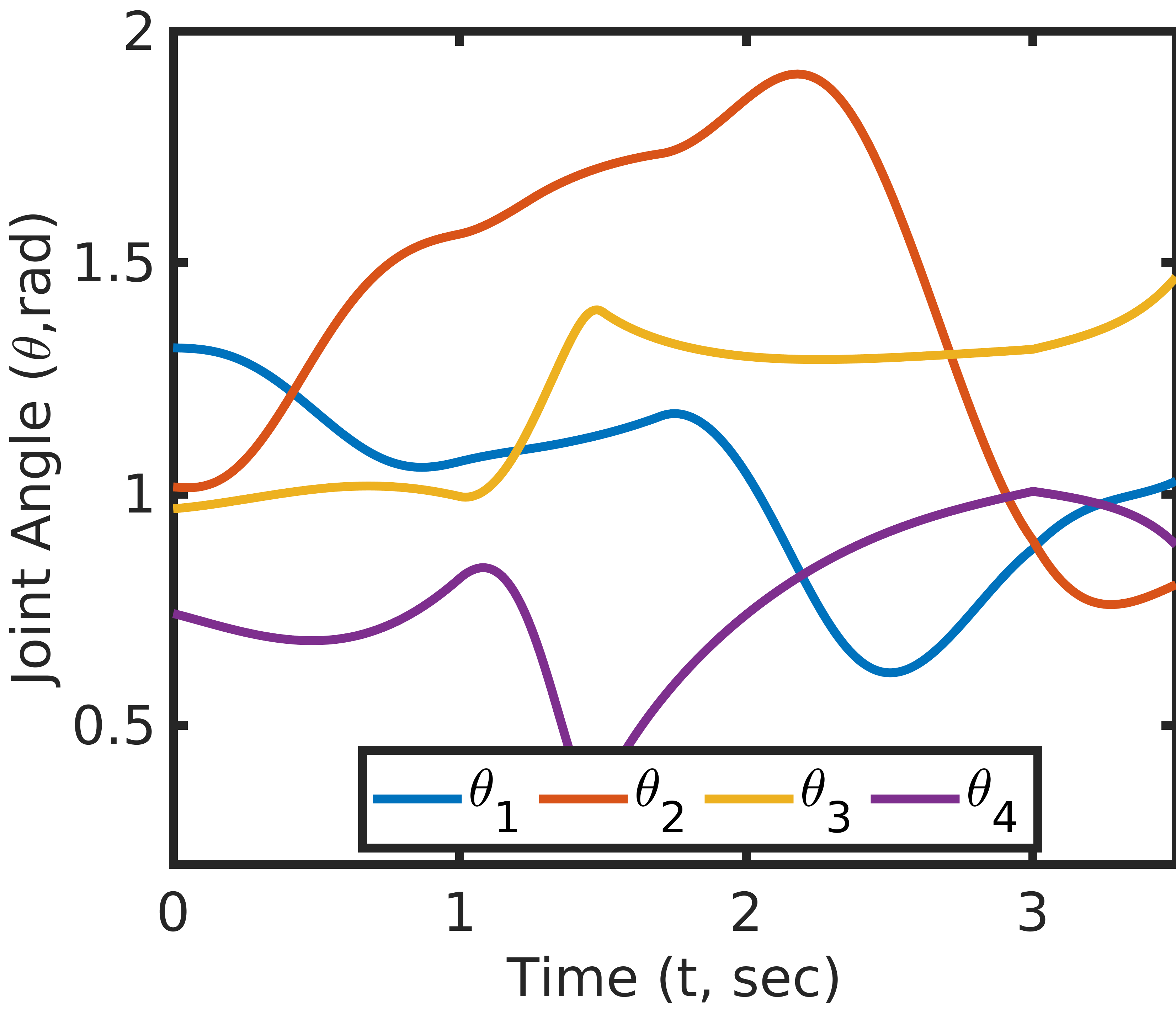}
\end{subfigure}%
\begin{subfigure}{.242\textwidth}
  \centering
  \includegraphics[width=0.95\linewidth]{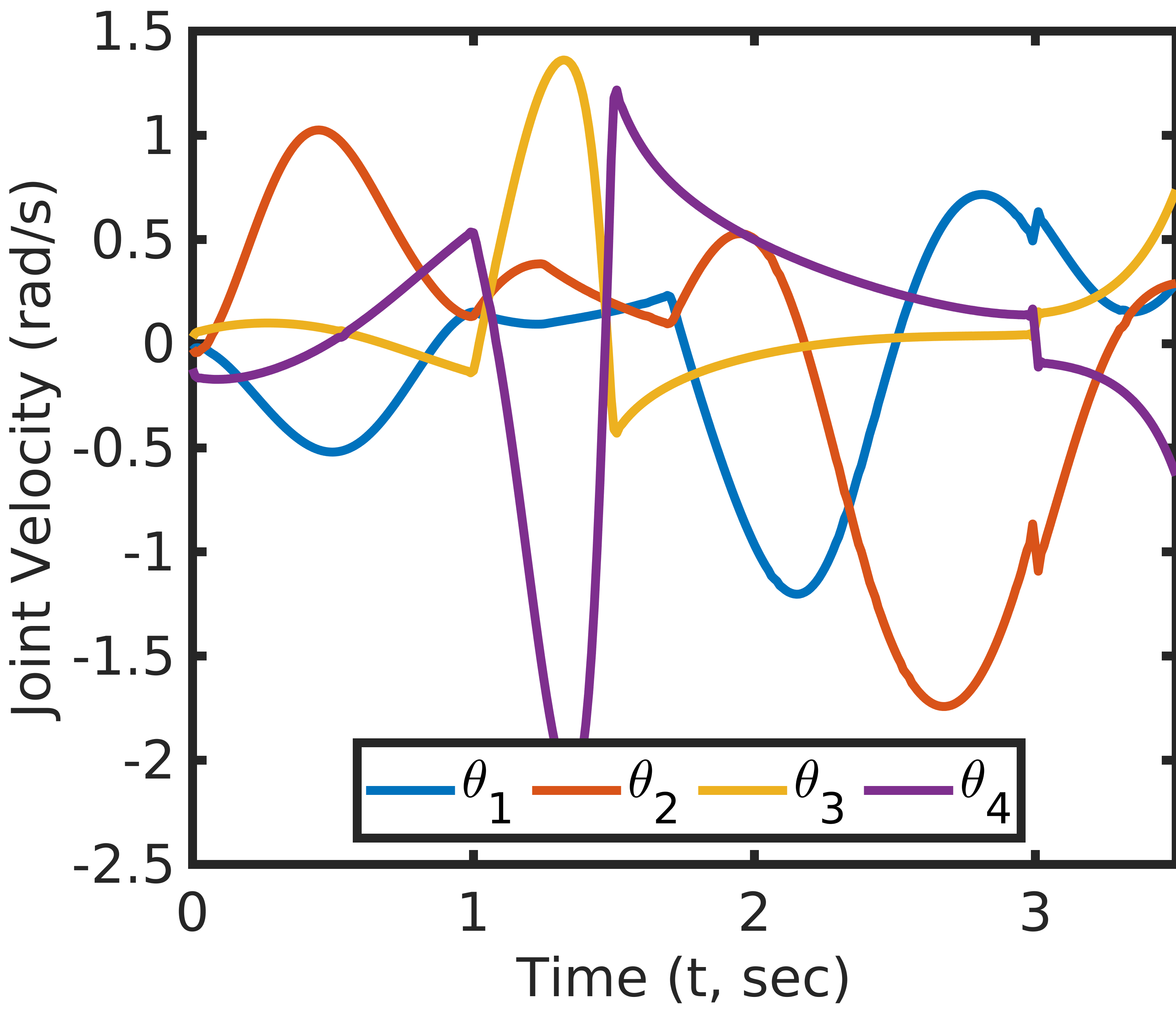}
\end{subfigure}
\begin{subfigure}{.242\textwidth}
  \centering
  \includegraphics[width=0.95\linewidth]{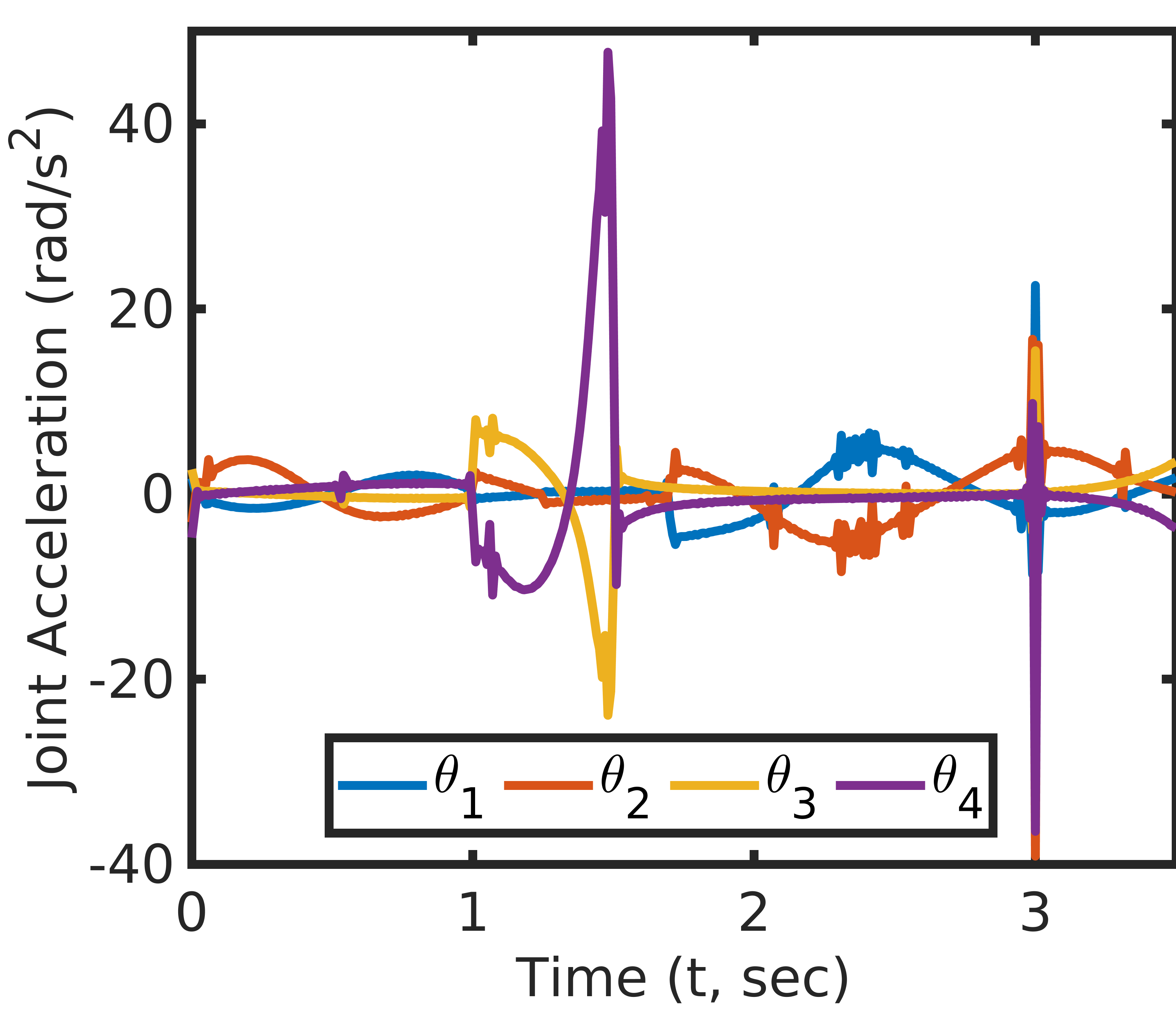}
\end{subfigure}
\begin{subfigure}{.242\textwidth}
  \centering
  \includegraphics[width=0.95\linewidth]{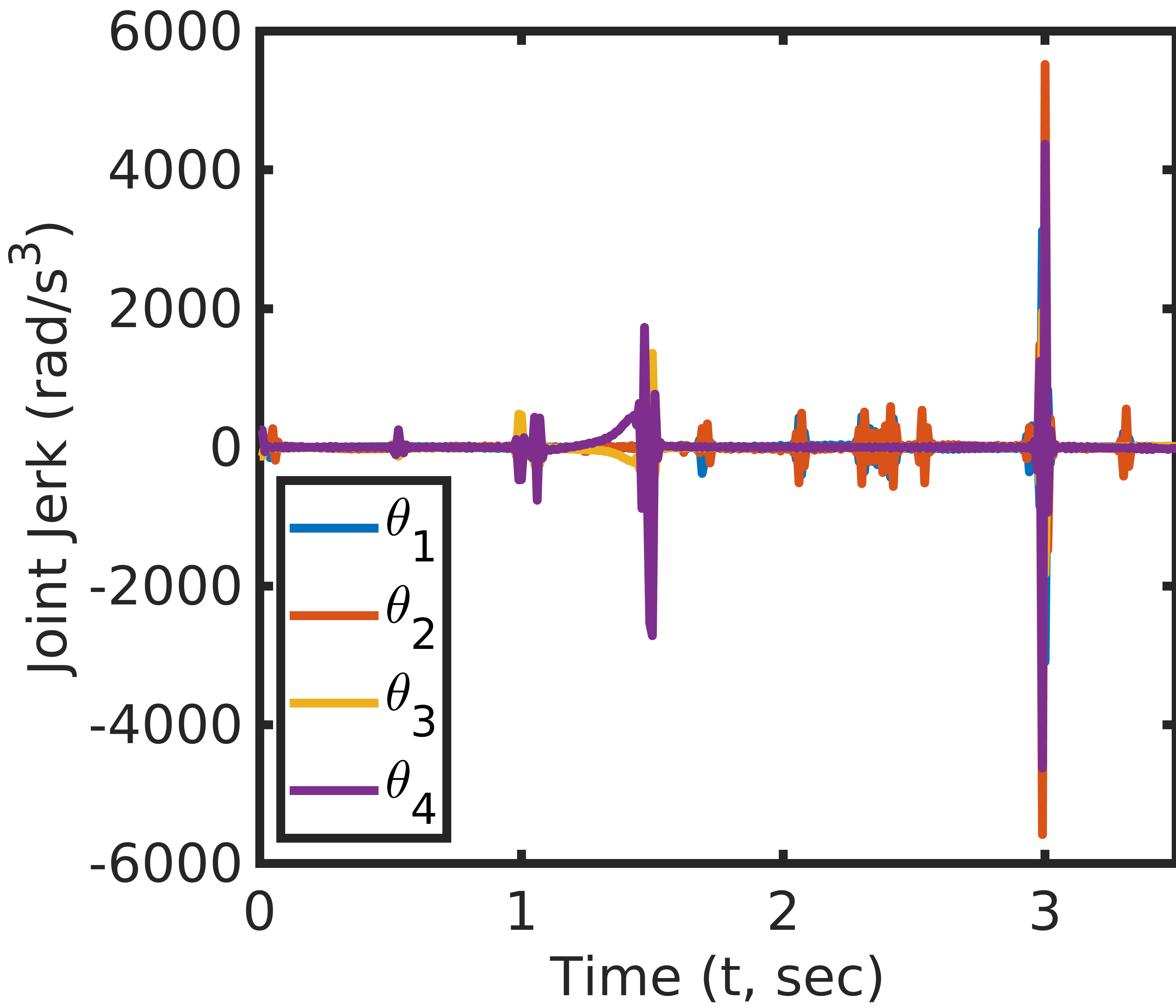}
\end{subfigure}
\caption{Joint Angle Position, Velocity, Acceleration and Jerk Plots. The first row corresponds to the brachistochrone hip trajectory, the second row for the circular arc trajectory and the third row for the virtual slope method}
\label{fig:posvel}
\end{figure*}

The overall simulated motion consists of the complete swing leg motion, landing DSP of the stance leg and the various hip trajectories, namely as a brachistochrone, as a circular arc and as a virtual slope. The commencement of DSP and then the swing phase and overall knee position based on the IKNN solutions can be referred from Fig. \ref{fig:swing}. The motion for the various hip trajectories are shown row wise in Fig. \ref{fig:motion}. Although the motion seems to be quite similar at the instants, but a close look at the angles creates the difference. The motion visualization and comparison in the joint space is shown cumulatively in Fig. \ref{fig:posvel}. The variation of the significant angles, $\theta_1, \theta_2, \theta_3$ and $\theta_4$ is considered as the other angles are either constant throughout the motion i.e. $\theta_5 = \pi/2$ or move in a very predefined manner i.e. $\theta_6, \theta_7, \theta_8$ and $\theta_9$. The difference between the accelerations and jerks at the knot points can be visualized and the values can be seen. A numerical comparison of significant factors can be seen in Table \ref{tablecomp}.
   
\begin{table}[h]
\caption{Comparison of Various Hip Traectories (Approx. Values)}
\label{tablecomp}
\centering
\begin{tabular}{|c|c|c|c|}
\hline
 & Brachistochrone & Circular & Virutal Slope\\
\hline
Max $z_{C_i}$ & 73 cm & 73 cm & 68 cm\\
\hline
Max abs. Acc. & 32 $rad/s^2$ & 38 $rad/s^2$ & 48 $rad/s^2$\\
\hline
Max abs. Jerk & 2000 $rad/s^3$ & 2700 $rad/s^3$ & 4700 $rad/s^3$\\
\hline
\end{tabular}
\end{table}

All the values of the initial hip position is based on the adaptive planning strategy to make the trajectory plan compatible for links with varying lengths and proportions. It can be observed that the value of initial hip position had to be changed for the linear case without changing the governing parameters of Table \ref{table:param}. This was because of the fact that linear hip motion was not able to satisfy the stance leg landing DSP workspace constraints with the workspace of the 2-link manipulator with base at hip.

\section{CONCLUSIONS}

An effective comparison of various hip trajectories during downstairs motion of a toe-foot biped robot model is established. The comparison along with the various motion instant visualizations show a more human like behavior. An adaptive initial hip height policy is developed and simulated along with a brachistochrone, a circular arc and a virtual slope based hip trajectory. Further works include proper torque analysis based on the dynamic model of the robot.

\end{document}